\documentclass[11pt,a4paper]{article}
\usepackage{fullpage}

\usepackage{graphicx} 
\usepackage{subfigure} 

\usepackage[round]{natbib} 
\usepackage{algorithm}
\usepackage{algorithmic}

\usepackage{hyperref}

\usepackage[T1]{fontenc}    
\usepackage{hyperref}       
\usepackage{url}            
\usepackage{booktabs}       
\usepackage{amsfonts}       
\usepackage{nicefrac}       
\usepackage{microtype}      
\usepackage{hyperref}
\usepackage{url}
\usepackage{amsmath,amssymb,amsthm,bm}
\usepackage{graphicx}
\usepackage{subfigure}
\usepackage{color}
\usepackage{pifont}
\usepackage{balance}
\usepackage{authblk}

\usepackage{balance}

\begin{document}
\title{Tensor Regression Meets Gaussian Processes}
\author[1]{Rose Yu \thanks{The work was performed while at USC.}}
\author[2]{Guangyu Li }
\author[2]{Yan Liu }
\affil[1]{Department of Computing and Mathematical Sciences, Caltech}
\affil[2]{Department of Computer Science, \ University of Southern California}
\date{}
\maketitle

\newcommand*\diff{\mathop{}\!\mathrm{d}}
\newcommand{\loss} {\mathcal{L}}
\newcommand{\W} {\mathcal{W}}

\newcommand\smallO{
	\mathchoice
	{{\scriptstyle\mathcal{O}}}
	{{\scriptstyle\mathcal{O}}}
	{{\scriptscriptstyle\mathcal{O}}}
	{\scalebox{.7}{$\scriptscriptstyle\mathcal{O}$}}
}

\newcommand{\R}{{\mathbb R}}   
\newcommand{\E}{{\mathbb E}}   

\newcommand{\skt}{{\mathcal{S}}}
\newcommand{\tr}{{\text{tr}}}
\newcommand{\vt}{{\text{vec}}}

\newcommand{\eat}[1]{}
\newtheorem{theorem}{Theorem}[section]
\newtheorem{proposition}[theorem]{Proposition}

\newtheorem{definition}[theorem]{Definition}
\newtheorem{lemma}[theorem]{Lemma}
\newtheorem{fact}[theorem]{Fact}
\newtheorem{corollary}[theorem]{Corollary}

\newenvironment{example}[1][Example]{\begin{trivlist}
\item[\hskip \labelsep {\bfseries #1}]}{\end{trivlist}}
\newenvironment{remark}[1][Remark]{\begin{trivlist}
\item[\hskip \labelsep {\bfseries #1}]}{\end{trivlist}}

\newcommand{\real}{\mathbb{R}}
\newcommand{\ex}{\mathbb{E}}
\newcommand{\pq}[1]{\left( #1 \right)}
\newcommand{\Loss}{\mathcal{L}}
\newcommand{\M}[1]{{\mathbf{#1}}} 
\newcommand{\T}[1]{{\mathcal{#1}}} 
\newcommand{\V}[1]{{\mathbf{#1}}} 
\newcommand{\SM}[1]{\mathbf{\hat{#1}}}
\newcommand{\TM}[1]{\mathbf{\bar{#1}}}
\newcommand{\krp}{\odot}
\newcommand{\PR}{\mathbb{P}}
\newcommand{\MG}[2]{\mathcal{N}(#1,#2)}
\newcommand{\OO}[1]{\mathcal{O}(#1)}
\newcommand{\OOt}[1]{\tilde{\mathcal{O}}(#1)}
\newcommand{\NI} [1]{ \N{ #1}_{1}}
\newcommand{\Nsp}[1]{ \N{ #1}_{2}}
\newcommand{\N}[1]{ {\left\| #1 \right\|}}
\newcommand{\Nf} [1]{ \N{ #1}_{\textsc{F}}}
\newcommand{\Mij}[3]{#1_{{#2},{#3}}}
\newcommand{\Mcij}[3]{{\{#1\}}_{{#2},{#3}}}
\newcommand{\SV}[2]{\sigma_{#1}(#2)}
\newcommand{\EV}[2]{\lambda_{#1}(#2)}
\newcommand{\numberthis}{\addtocounter{equation}{1}\tag{\theequation}}
\newcommand{\abs}[1]{ {\left| #1 \right|}}

\newcommand{\ryedit}[1]{{\color{magenta} #1}}
\newcommand{\rycomment}[1]{\ryedit{[RY: #1]}}
\newcommand{\dhedit}[1]{{\color{red} #1}}
\newcommand{\dhcomment}[1]{\dhedit{[DH: #1]}}
\newcommand{\yledit}[1]{{\color{blue} #1}}
\newcommand{\ylcomment}[1]{\yledit{[YL: #1]}}

\newcommand{\kp}{\otimes}
\newcommand{\hp}{*}
\newcommand{\op}{\circ }
\newcommand{\cpd}[1]{{\left\llbracket {#1} \right\rrbracket}}

\newcommand{\fix}{\marginpar{FIX}}
\newcommand{\new}{\marginpar{NEW}}
	
	\begin{abstract}
Low-rank tensor regression, a new model class that learns high-order correlation from data, has  recently received considerable attention. At the same time, Gaussian processes (GP) are well-studied machine learning models for structure learning.  In this paper,  we demonstrate interesting  connections between  the two, especially for multi-way data analysis. We show that low-rank tensor regression is  essentially learning a multi-linear kernel in Gaussian processes, and the low-rank assumption translates to the constrained Bayesian inference problem.  We prove the oracle inequality and derive the average case learning curve for the equivalent GP model. Our finding implies that low-rank tensor regression,  though empirically successful,  is highly dependent on the eigenvalues of covariance functions as well as  variable correlations. 
		
%
	\end{abstract}
	
	\section{Introduction}
High-order correlations are ubiquitous in modern data analytics.  For instance, data generated from a sensor network contain measurements from different locations, time stamps, and variables. Accurate prediction requires models that can simultaneously capture correlations  across time, space and variables.  Low-rank tensor regression is a class of supervised learning models that aim to learn such high-order correlations. In recent years, low-rank  tensor regression  has been intensively studied in machine learning, leading to successful applications in multi-task learning \citep{wimalawarne2014multitask},  deep learning \citep{novikov2015tensorizing},  complex network analysis \citep{imaizumi2016doubly}.


In contrast to  traditional unsupervised tensor decomposition \citep{kolda2009tensor}, tensor regression \citep{zhou2013tensor} learns a tensor model in a supervised fashion and imposes low-rank structure for dimension reduction. Tensor regression has several advantages over vector or matrix regression: from  the modeling perspective,  the model tensor  provides an explicit parameterization for  the  multi-directional interdependence among variables. The  low-rankness represents the shared latent space in the data.  From the learning perspective, tensor model enjoys lower sample complexity. The tensor low-rank constraint regularizes the model to be more generalizable.  However, a notable disadvantage of tensor regression is the absence of confidence intervals for the predictions, which calls for a  probabilistic counterpart that can effectively represent the high-order correlations in the data.

Meanwhile, Gaussian processes  \citep{rasmussen2006gaussian} are  well-established techniques for modeling correlations  structures. With versatile covariance design, GP remain popular  in  spatial statistics and time series analysis. A natural question then arises, ``which method is better? And how are these two model classes related?'' Known examples of  similar connections include the Gaussian process latent variable model   \citep{lawrence2004gaussian} for PCA,   the multi-task Gaussian process model \citep{bonilla2007multi} for multi-task learning and the probabilistic Tucker model for Tucker tensor decomposition \citep{chu2009probabilistic}.  The probabilistic interpretation deepens the understanding of the regularized optimization approach, suggesting its generalization to non-Gaussian  data with kernel methods.



In this paper, we make the first attempt at understanding this connection.  We show that tensor regression is equivalent to learning a Gaussian process with multi-linear transformation kernel: multi-linear Gaussian process (MLGP). The low-rank assumption on the parameter tensor can be  interpreted as a constrained Bayesian inference problem. We analyze the theoretical properties of MLGP by proving its oracle inequality  and deriving the average case learning curve.  We validate our theory with numerical simulations  and provide a comparative analysis between different GP models. Finally, we showcase the model on three real-world tensor regression applications: multi-linear multi-task learning, spatio-temporal forecasting, and multi-output regression. The model not  only can achieve superior  performance  but also uncover  interesting patterns from multi-way data.

Note that the goal of our work is fundamentally different from existing works on Bayesian estimator for tensor-variate regression  \citep{guhaniyogi2015bayesian, xu2015bayesian,suzuki2015convergence}.  For example, \citep{xu2015bayesian} propose a generative model for Bayesian tensor regression; \citep{suzuki2015convergence}  analyzes the minimax optimal rate of the estimator.  These works  emphasize probabilistic modeling instead of establishing the connections.  And most existing theoretical analyses are asymptotic. In contrast, our work aims to provide deeper insights into the relationship between the optimizers of tensor regression  and  estimators for  Gaussian process models.





	\vspace{-2mm}
    
	
	
	\section{Tensor Regression and Its Counterpart}
	\subsection{ Low-Rank Tensor Regression}

Tensor regression exploits the high-order correlation in the data. It learns a multi-linear function whose parameters form a  tensor. To represent shared latent spaces and address ``the curse of dimensionality'' issue,   tensor regression usually constrains the mode tensor to be low-rank. Formally, given an input tensor $\T{X} $, an output tensor $\T{Y} $ and a model parameter tensor $\W$,  tensor regression aims to solve the following optimization problem: 
  \begin{eqnarray}
 	\label{eqn:tensor_regression}  \nonumber
 	\W^\star = \text{argmin}_{\W} \hat{\loss}( f(\T{X}, \W); \T{Y})  
\\	\text{s.t.} \quad  \text{rank} (\W) \leq R 
 \end{eqnarray}
where  $\hat{\loss}$ denotes the loss function, and $f$ represents a regression model (e.g. linear, logistic). The solution $\T{W}^\star$ minimizes the empirical loss ,  subject to the tensor low-rank constraint $\text{rank} (\W) \leq R $. 

Low-rank tensor regression has many applications. One example is multi-linear multi-task learning \footnote{Other applications can be re-formulated as special cases \\ of multi-linear multi-task learning} (MLMTL), which  learns multiple tasks with a multi-level task hierarchy. For example, when  forecasting the energy demand for multiple power plants, we can split the  tasks by categories: coal, oil and natural gas.  MLMTL improves the prediction  by modeling the correlations within and across categories. We can encode such  task hierarchy using a tensor, where the first dimension of the tensor represents features, and the rest to index the grouped tasks at each level.
 
Specifically, given $T$ learning tasks  with feature dimension $T_1$, we can split them into $T_2$ groups, each of which contains $T_3=T/T_2$ tasks. Assuming each task $t$ contains $n_t$ training data points $ \{\V{x}_{t,i}, \V{y}_{t,i} \}_{i=1}^{n_t}$ and is parametrized by $\V{w}_t \in \R^{T_1}$.  We can form a  tensor by concatenating all the parameters as a matrix $ \M{W} = [\V{w}_1,\cdots, \V{w}_T]$ and folding along the feature dimension $\W = \text{fold}_{(1)}(\M{W}) \in \R^{T_1 \times T_2 \times T_3}$. The objective of MLMTL is to learn this parameter tensor subject to the low-rank constraint: 
\begin{eqnarray}
  	\W^\star & = \text{argmin}_{\W} 	\sum_{t =1}^T \sum_{i=1}^{n_t} \loss (\langle \V{x}_{t,i}, \V{w}_t \rangle  ; \V{y}_{t,i} ) \nonumber \\
  	 \text{s.t.}& \quad  \text{rank}(\T{W} )\leq R
 	\label{eqn:multitask}
 \end{eqnarray}
 
If  the task hierarchy has two levels  $T = T_2 \times T_3$, we obtain a third-order tensor.  In general, one can use an $(m+1)$-order tensor to represent an $m$-level task clustering hierarchy. Note that the definition of tensor rank is not unique \citep{kolda2009tensor}. One popular  definition is Tucker rank due to its computational benefit.  Tucker rank assumes that the tensor $\T{W}$ has a Tucker decomposition $\W = \T{S} \times_1 \M{U}_1 \times_2 \M{U}_2\times_3\M{U}_3$, with a core tensor $\T{S}\in \R^{R_1 \times R_2 \times R_3}$ and orthonormal projection matrices $\{\M{U}_m \}_{m=1}^3$. Tucker rank corresponds to the size of the core tensor $\T{S}$.


Low-rank tensor regression  is a challenging problem mainly due to  the subspace of low-rank tensors is non-convex, resulting in a high-dimensional non-convex problem.  Recent developments have seen efficient algorithms for  solving  Equation \ref{eqn:tensor_regression} and \ref{eqn:multitask}, e.g., \citep{yu2016learning, rabusseau2016low}, demonstrating low-rank tensor regression as a scalable method for multi-way data analysis. However,  one major drawback of such formulation is that it trades  uncertainty for efficiency: there is no confidence interval for the prediction.  Hence, it is difficult for the learned tensor model to reason with uncertainty. In seek of its probabilistic counterpart, we resort to another class of structured learning models: Gaussian processes.

\subsection{Multi-linear Gaussian Processes}

Gaussian process regression infers continuous values with a GP prior. Given input $\V{x}$, output $\V{y}$, and a regression model
\begin{equation}
\V{y} = f(\V{x}) + \epsilon, \quad f(\V{x}) \sim \text{GP}(m,k) 
\end{equation}
with  $\epsilon$ as the Gaussian noise.  GP characterizes a prior distribution over function $f(\V{x})$ with a mean function $m$ and a covariance function $k$. By definition, we have $\E[f(\V{x})]= m(\V{x})$, $\text{cov} (\V{x}, \V{x}') = k(\V{x}, \V{x}')$. The mean function is usually defined to be zero. The covariance function completely defines the process's behavior.

Next, we develop a  GP model to describe the  generative process of the MLMTL problem. Given a total of $N = \sum_{t=1}^T n_t$ training data points$ \{\V{x}_{t,i}, \V{y}_{t,i} \}_{i=1}^{n_t}$ from $T$ related  tasks,  we assume that each data point $(\V{x}_{t,i}, \V{y}_{t,i})$ is drawn i.i.d  from the following probabilistic model:
\begin{equation}
\V{y}_{t,i} = f(\V{x}_{t,i}) +  \epsilon_t, \quad  f(\V{x}_{t,i}) \sim \text{GP}(0,k) 
\label{eqn:mlgp}	
\end{equation}
where the task $t$ has a Gaussian noise $\epsilon_t \sim N(0, \sigma_t^2)$ with zero mean and variance $\sigma_t^2$. To model multiple tasks,  we can concatenate the data from all tasks:
\begin{align*}
\V{y} = \begin{bmatrix}
 \V{y}_{t,1}\\ 
  \V{y}_{t,2}\\ 
 \cdots \\
 \V{y}_{T,n_T}
	\end{bmatrix}, 
   	\quad 
	\M{X} = \begin{bmatrix}
		\M{X}_{1} & \V{0} & \dots  &\V{0}  \\
		\V{0} & \M{X}_{2}   & \dots  & 	\V{0}  \\
		\vdots & \vdots & \ddots & \vdots \\
	\V{0}  & \V{0}  & \dots  & \M{X}_{T}
	\end{bmatrix},
			\M{D} = \begin{bmatrix}
			\sigma_1^2 \otimes I_{n_1} & \V{0}  & \dots  &\V{0}  \\
			\V{0} & \sigma_2^2\otimes I_{n_2}   & \dots  & 	\V{0}  \\
			\vdots & \vdots & \vdots & \vdots \\
			\V{0}  & \V{0}  & \dots  & \sigma_T^2 \otimes I_{n_T}
			\end{bmatrix}
\end{align*}
\nonumber 
where $\M{X}_{t} = [\V{x}_{t,1} ;\V{x}_{t,2} ;\cdots; \V{x}_{t,n_t} ]$ is the vectorization of the inputs for task $t$. In matrix form, the probabilistic model  generalizes  Equation \ref{eqn:mlgp} into:
\begin{equation*}
	\V{y} = f(\M{X}) +  \M{e},\quad f(\M{X})\sim  \text{GP}(\V{0}, \M{K}), \quad \M{e} \sim \T{N}(\M{0}, \M{D})
\end{equation*}
 with   $\M{X}$ as the inputs, $\M{K}$ as the input covariance matrix  and $\M{D}$ as the  noise covariance.  

\begin{figure}[htbp]
    \centering
    	\subfigure[\textsf{LIN}]{
        \includegraphics[width=0.2\columnwidth]{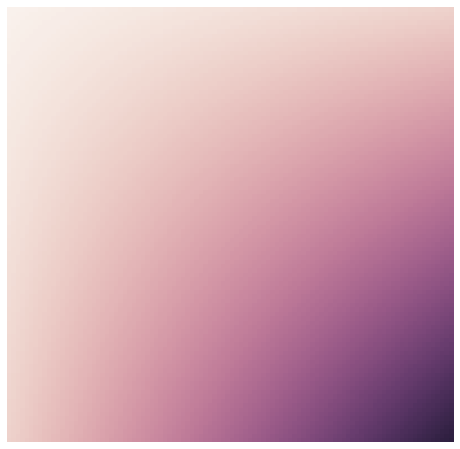}
		}
        \subfigure[\textsf{EXP}$\otimes$\textsf{LIN}]{
        \includegraphics[width=0.2\columnwidth]{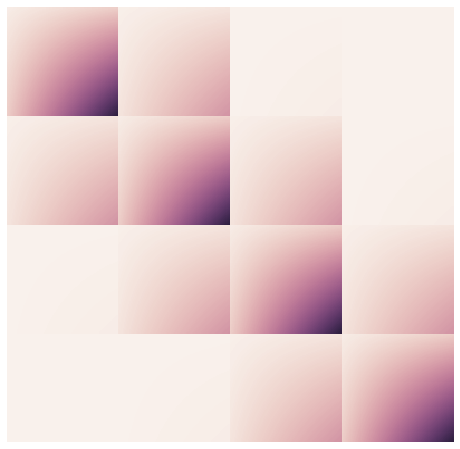}
		}
  \subfigure[\textsf{PED}$\otimes$\textsf{EXP}$\otimes$\textsf{LIN}]{
        \includegraphics[width=0.2\columnwidth]{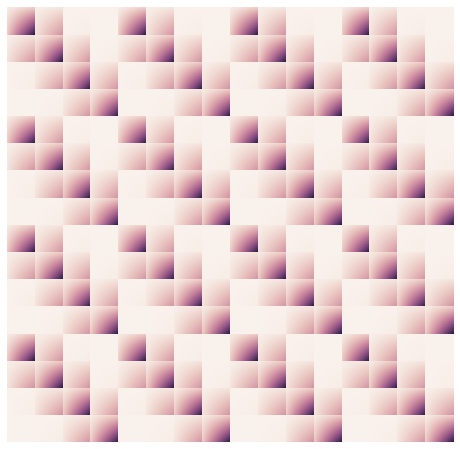}
		}\\
       
       \subfigure[\textsc{EXP}]{
        \includegraphics[width=0.2\columnwidth]{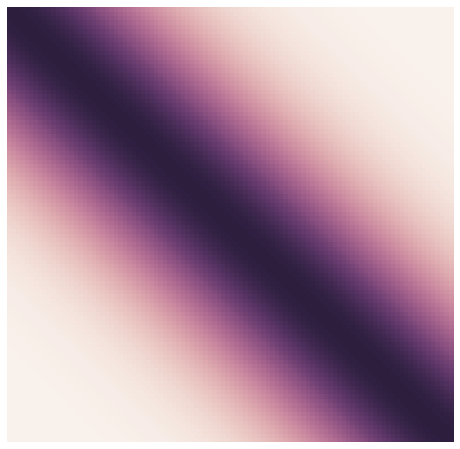}
		}
        \subfigure[\textsf{LIN}$\otimes$\textsf{EXP}]{
        \includegraphics[width=0.2\columnwidth]{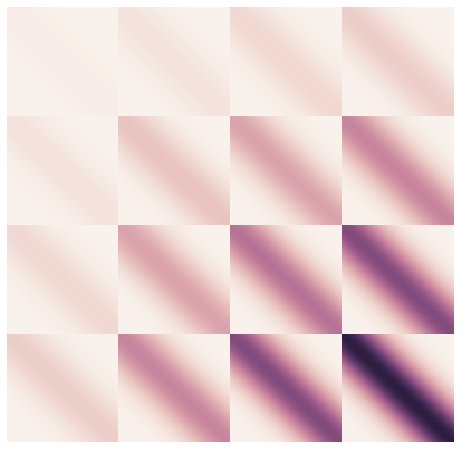}
		}
        \subfigure[\textsf{PED}$\otimes$\textsf{LIN}$\otimes$\textsf{EXP}]{
        \includegraphics[width=0.2\columnwidth]{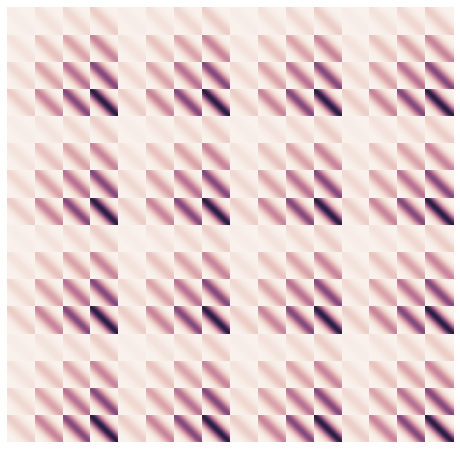}
		} \\
       
        \subfigure[\textsc{PED}]{
        \includegraphics[width=0.2\columnwidth]{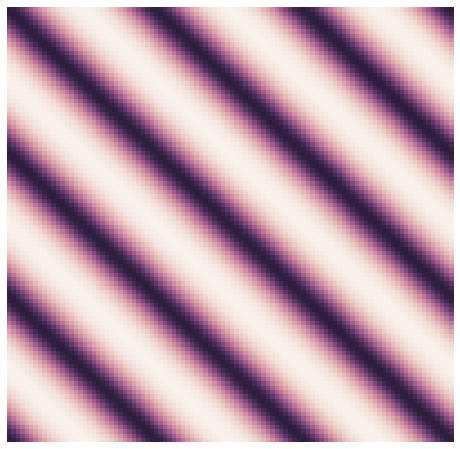}
		}
        \subfigure[\textsf{EXP}$\otimes$\textsf{PED}]{
        \includegraphics[width=0.2\columnwidth]{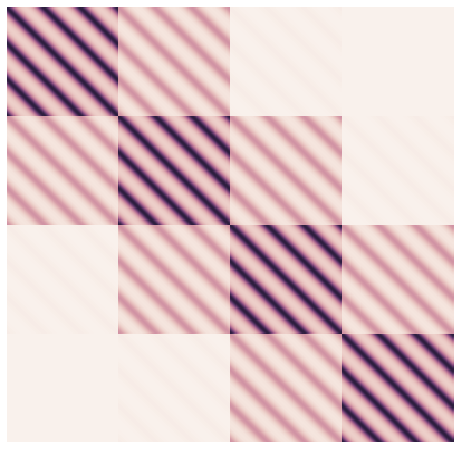}
		}
   \subfigure[\textsf{LIN}$\otimes$\textsf{EXP}$\otimes$\textsf{PED}]{
        \includegraphics[width=0.2\columnwidth]{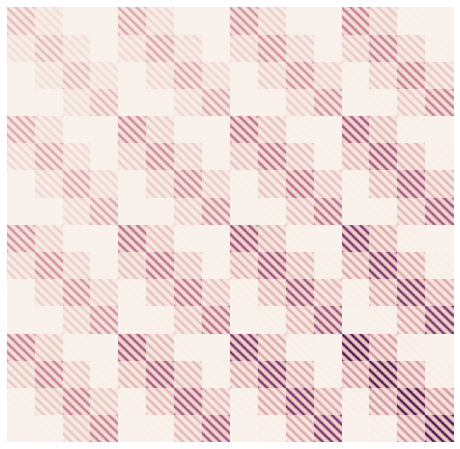}
		} 
   \caption{Visualization of the multi-linear kernel,  constructed by iteratively  composing Linear (\textsf{LIN}) Squared Exponential  (\textsf{EXP}) and Period  (\textsf{PED}) kernels on a $(50\times 50) \otimes (4 \times 4)  \otimes (4\times 4) $ grid following different order.}\label{fig:multi-linear kernel}. 
\end{figure}

To represent the multi-level task hierarchy $T=T_2\times T_3$, we define the  kernel matrix $\M{K}$ with Kronecker products:
\begin{eqnarray*}
\M{K} = \phi(\M{X}) \M{K}_3 \otimes \M{K}_2 \otimes  \M{K}_1\phi(\M{X})^\top
\end{eqnarray*}
where $\M{K}_1$ models the feature correlations,  $\M{K}_2$ models the correlations across groups, and $\M{K}_3$ represents the dependences of tasks within the group.  $\phi(\cdot) $ maps the inputs to a $T_1$ dimensional feature space. \footnote{We want to clarify that the use of $\phi(\cdot)$ limits the model to a finite feature space. And the model itself is parametric, which is the same as the tensor regression formulation.} This multi-linear kernel provides a multi-resolution compositional representation. It is expressive yet efficient. Figure \ref{fig:multi-linear kernel} shows several examples of such construction with three  kernel functions: Linear $k(x,x')=a+b(x-c)(x'-c)$, Squared Exponential $k(x,x')=a \exp{\frac{-(x-x')^2}{2c}}$ and Periodic $k(x,x') = a \exp{-\frac{sin^2(\pi |x-x'|)}{c}}$ in different orders.   We name this class of GP model  multi-linear Gaussian processes (MLGP) as the kernel matrix encodes multi-linear structure. 

\subsection{Connection Between Two Models}
In the following section, we connect low-rank tensor regression with multi-linear Gaussian processes by examining the common structures that the two models aim to learn.

\begin{figure*}[htbp]
	\begin{center}
		\subfigure[Tensor Regression]{
			\includegraphics[scale = 0.26]{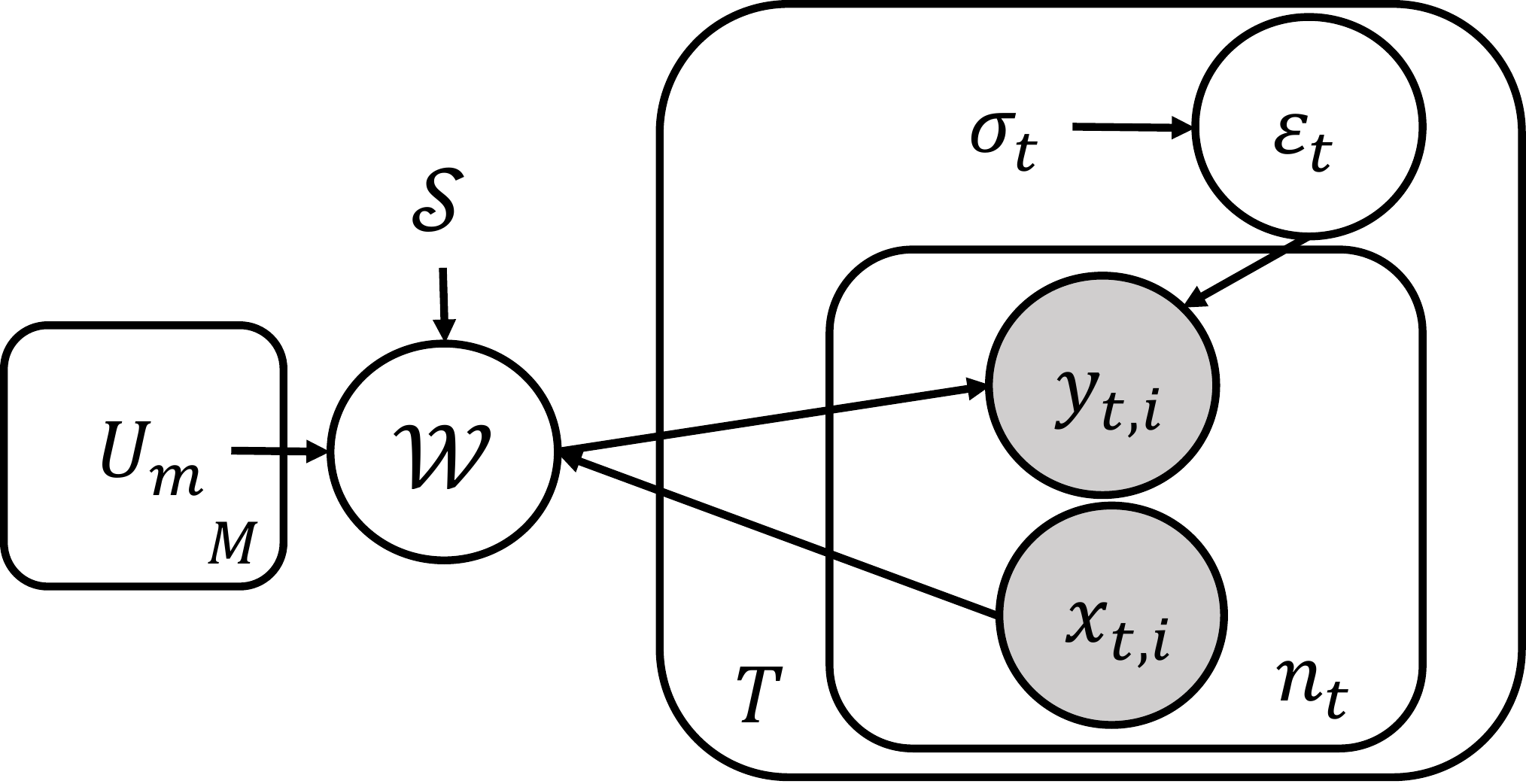}
			\label{fig:graph-tr}
		}
		\subfigure[Gaussian Process]{
			\includegraphics[scale = 0.26]{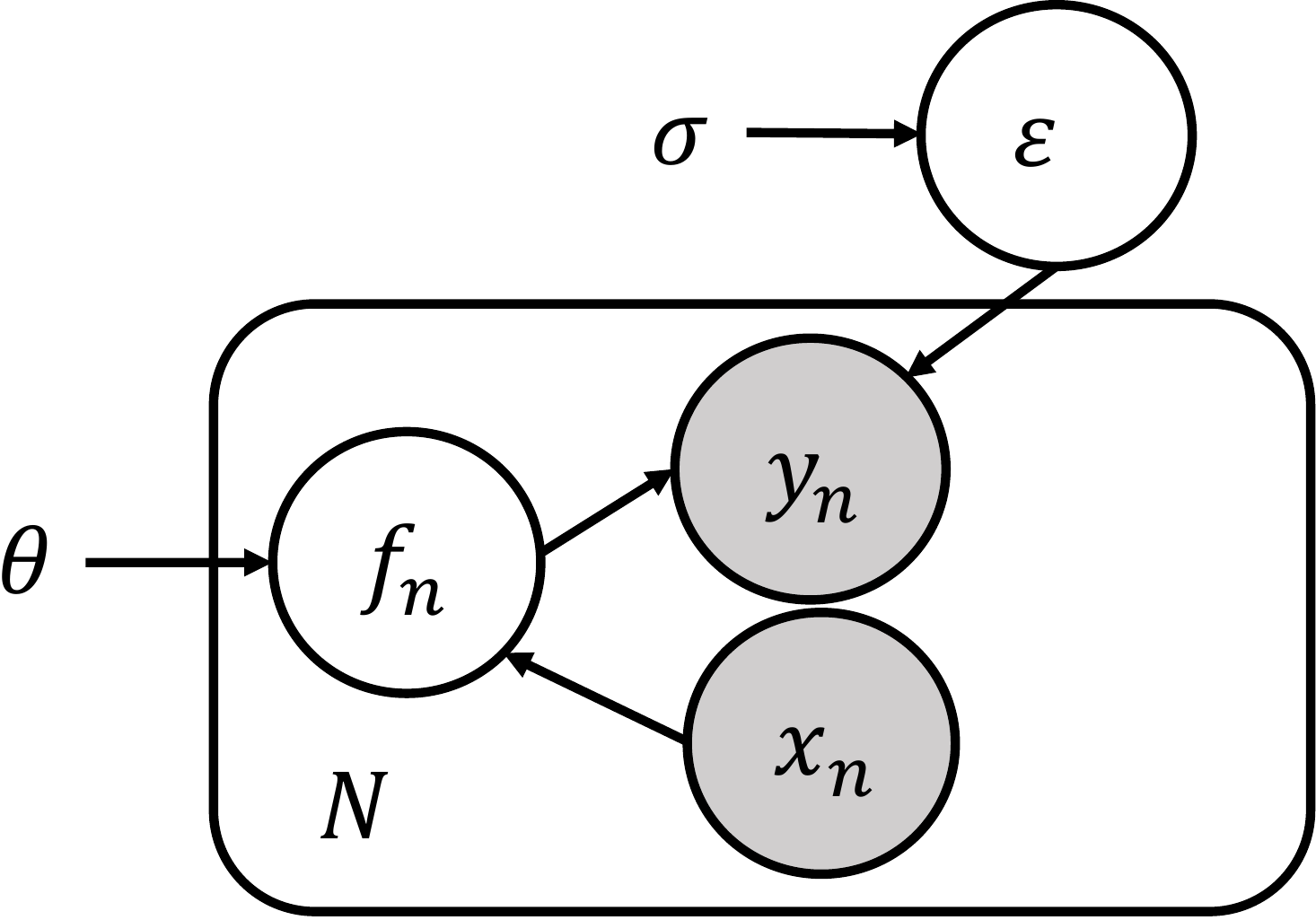}
			\label{fig:graph-gp}
		}
		\subfigure[Multi-linear Gaussian Process] {
			\includegraphics[scale = 0.26]{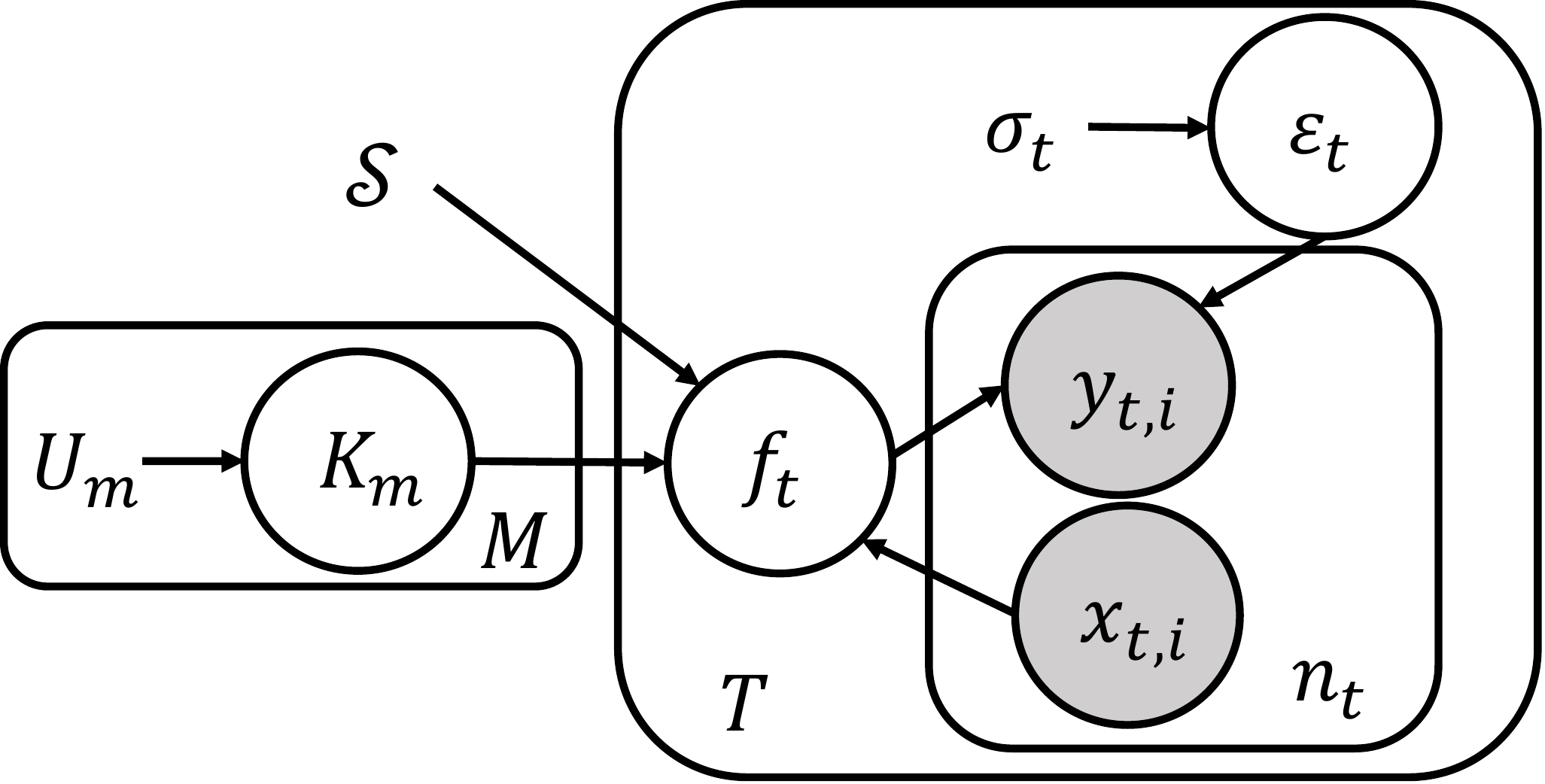}
			\label{fig:graph-mlgp}
		}
	\end{center}
		\vspace{-5mm}
	\caption{Graphical model  for \subref{fig:graph-tr} tensor regression, \subref{fig:graph-gp} Gaussian process and  \subref{fig:graph-mlgp} MLGP.  The outer plate represents tasks, while the inner plate represents the repeated examples
	within a task.}
	\label{fig:graphical_model}
\end{figure*}

When dealing with a large number of tasks and high dimensional data, learning $\{\M{K}_m\}_{m=1}^3$ can be very expensive. To reduce the computational cost, we use the low-rank approximation for each correlation matrix: 
\begin{eqnarray*}
\{\M{K}_{m} =& \M{U}_m \M{U}_m^\top \in \R^{T_m\times T_m } \}_{m=1}^3
 \label{eqn:gp_cov}
\end{eqnarray*}
where $\M{U}_m \in \R^{T_m \times R_m}$ is an orthogonal matrix with dimension $R_m$  much smaller than $T_m$. 

The weight-space view of GP allows us to re-write the latent function:  $f(\M{X}) = \langle 	\text{vec}(\T{W}) , \phi( \M{X}) \rangle$, where $\T{W}  \in \R^{T_1  \times T_2\times T_3}$ is the regression model parameters with the following prior distribution:
\begin{eqnarray*}
	\text{vec}(\T{W}) = (\M{U}_1 \otimes \M{U}_2  \otimes \M{U}_3 )^T \text{vec}(\T{S}) \quad
    \text{vec}(\T{S}) \sim \T{N} (\V{0}, \sigma^2_s\M{I}) 
	\end{eqnarray*}
Here $\T{S} \in \R^{R_1\times R_2 \times R_3}$ is a super-diagonal core tensor with i.i.d entries.  $\{\M{U}_m \in \R^{T_m \times R_m}\}$ is a set of orthogonal projection matrices.  

Under the MLGP model, the prior distribution of the latent function follows  Gaussian $p(f|\M{X}) = \T{N} (0,\M{K})$, and the likelihood distribution is $p(\V{y}|f ) = \T{N}(f, \M{D})$. By integrating out the model parameters, we can obtain the marginal distribution of the outputs $\V{y}$  :
\begin{eqnarray*}
	p(\V{y}|\V{X}) = & \int_{f} p(\V{y}, f, s| \V{X}) \diff f  \diff s=   \T{N}(\M{0},\M{K} + \M{D}) 
\end{eqnarray*}
where we omit the  core tensor constant $\sigma_s$, which acts as a regularization term.
The log-likelihood of the marginal distribution for MLGP  is:
\begin{eqnarray}
	L &=  -\frac{1}{2}\log  |\M{K}+ \M{D}|  -\frac{1}{2} \V{y}^\top (\M{K}+ \M{D})^{-1} \V{y} + const \nonumber \\
	  &\text{s.t.} \quad \M{K} = \phi(\M{X})\otimes_{m=1}^3 \M{K}_m \phi(\M{X})^\top  
\label{eqn:mle}
\end{eqnarray}
 Using the Kronecker product property $\otimes_{m=1}^3 \M{U}_m \M{U}_m^\top  =(\otimes_{m=1}^3 \M{U}_m )(\otimes_{m=1}^3 \M{U}_m )^\top$, we can re-write the covariance matrix as:
\begin{eqnarray*}
\M{K} = (\phi(\M{X})\otimes_{m=1}^3 \M{U}_m ) (\phi(\M{X})\otimes_{m=1}^3 \M{U}_m )^\top 
\end{eqnarray*}

Denote $\tilde{\M{U}}  = \phi(\M{X})\otimes_{m=1}^3 \M{U}_m$ and let the singular value decomposition of $\tilde{\M{U}}$ be $\tilde{\M{U}}  = \M{U}_x \M{\Sigma}_x \M{V}_x^\top$. We can maximize the log-likelihood  by taking  derivatives over  $L$ with respect to   $\tilde{\M{U}}$ and set it to zero, which gives the stationary point condition:
\begin{equation*}
\V{y}\V{y}^\top  (\M{K}+\M{D})^{-1} \tilde{\M{U}} = \tilde{\M{U}}
\label{eqn:stationary_point}
	\end{equation*}
With some manipulation, we can obtain an equivalent eigenvalue problem. Detailed derivation can be found in  Appendix \ref{sec:derive_eigen}.
\begin{equation*}
\V{y}\V{y}^\top \M{U}_x = \M{U}_x (\M{\Sigma}_x^2  + \M{D} )
\end{equation*}

Further perform  eigen-decomposition of the output covariance $\V{y}\V{y}^\top = \M{U}_y\M{\Lambda}_y\M{U}^{-1}_y$, we have $\M{U}_x = \M{U}_y$, $\M{\Sigma}_x = (\M{\Lambda}_y-\M{D})^{\frac{1}{2}}$. 
Therefore, the likelihood of the MLGP model is maximized when the solution satisfies 
\begin{equation}
\phi(\M{X})\otimes_{m=1}^3 \M{U}_m=  \M{U}_y (\M{\Lambda}_y-\M{D})^{\frac{1}{2}}  \M{V}_x^\top
\label{eqn:equivalence}
\end{equation}
which suggests that the  maximum likelihood estimator of MLGP correspond to  a multi-linear  transformation from  the feature space $\phi(\M{X})$ to the principal subspace of the output.  Recall that for tensor regression in Equation \ref{eqn:tensor_regression}, the model parameter tensor $\T{W}$ also maps  features to the output space  with principal subspace projection using the Tucker decomposition of $\T{W}$. Hence MLGP and tensor regression are essentially learning the same latent feature representations. 

If we further consider the low-rank structure in the projection matrices,   GP becomes degenerate. Degenerate GP has been shown in \citep{quinonero2005analysis} to be equivalent to finite sparse linear models. Alternatively, we can interpret the low-rankness  in  MLPG and tensor regression   using a constrained Bayesian inference approach \citep{koyejo2013constrained}. By minimizing  the Kullback-Leibler (KL) divergence  of the Bayesian posterior $\T{N}(\V{0}, \M{K} +\M{D})$ from any constructed  GP prior $\T{N}(\V{0}, \M{S})$, and assuming $\M{K}$ is low-rank, we have the following  problem:
\begin{align*}
\min_{\M{\M{K}}: \M{K} \succeq \M{0}, \text{rank}(\M{K}) < R}   \log \det [(\M{K} +\M{D})\M{S}^{-1}] +  	\text{tr}[(\M{K} +\M{D})^{-1}\M{S}] 
	\end{align*} 

It turns out that the log-det of $\M{K} +\M{D}$ is a \textit{smooth surrogate} for the rank  of $\M{K}$, which simultaneously minimizes the rank of  $\W$. Therefore, the estimator for MLGP with low-rank kernel provides an approximate solution to  the low-rank tensor regression problem.  To this end, we have established the connections between tensor regression and Gaussian processes. Figure \ref{fig:graphical_model} depicts the graphical models of tensor regression, GP, and MLGP.  It is evident that the parameter tensor in tensor regression maps to the  covariance of the MLGP model. Latent tensor components become parameters of the covariance function.


We employ  gradient-based optimization for Equation \ref{eqn:mle} to learn the hyper-parameters of MLGP. (see Appendix \ref{sec:derive-opt} for details) Note that  gradient-based optimization does not guarantee the orthonormality of the projection matrices.  However, with a good initialization, we can still obtain reasonable approximations. As $\M{K}$ contains the Kronecker product and the low-rank structure, we can apply Woodbury matrix identity  and exploit Kronecker properties  to  speedup  the  inference.  The predictive distribution for the test data  follows  the standard GP regression procedure and has a closed form solution. 


	\subsection{Theoretical Analysis}

\begin{figure*}[t]
	\begin{center}
		\subfigure[Full-rank MLGP]{
			\includegraphics[scale = 0.20]{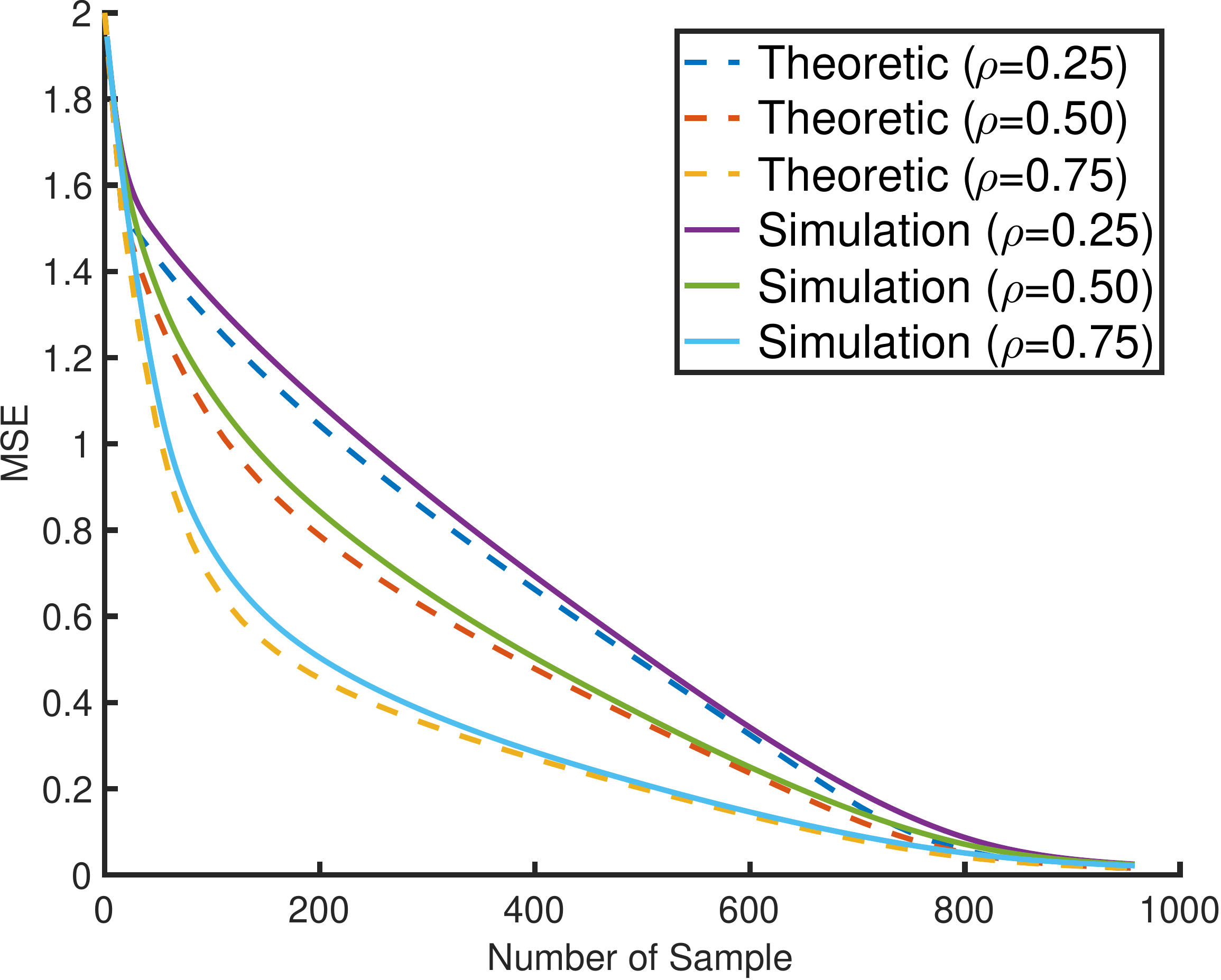}
			\label{fig:theory-full}
		}
		\subfigure[Low-rank $2$-mode MLGP] {
			\includegraphics[scale = 0.19]{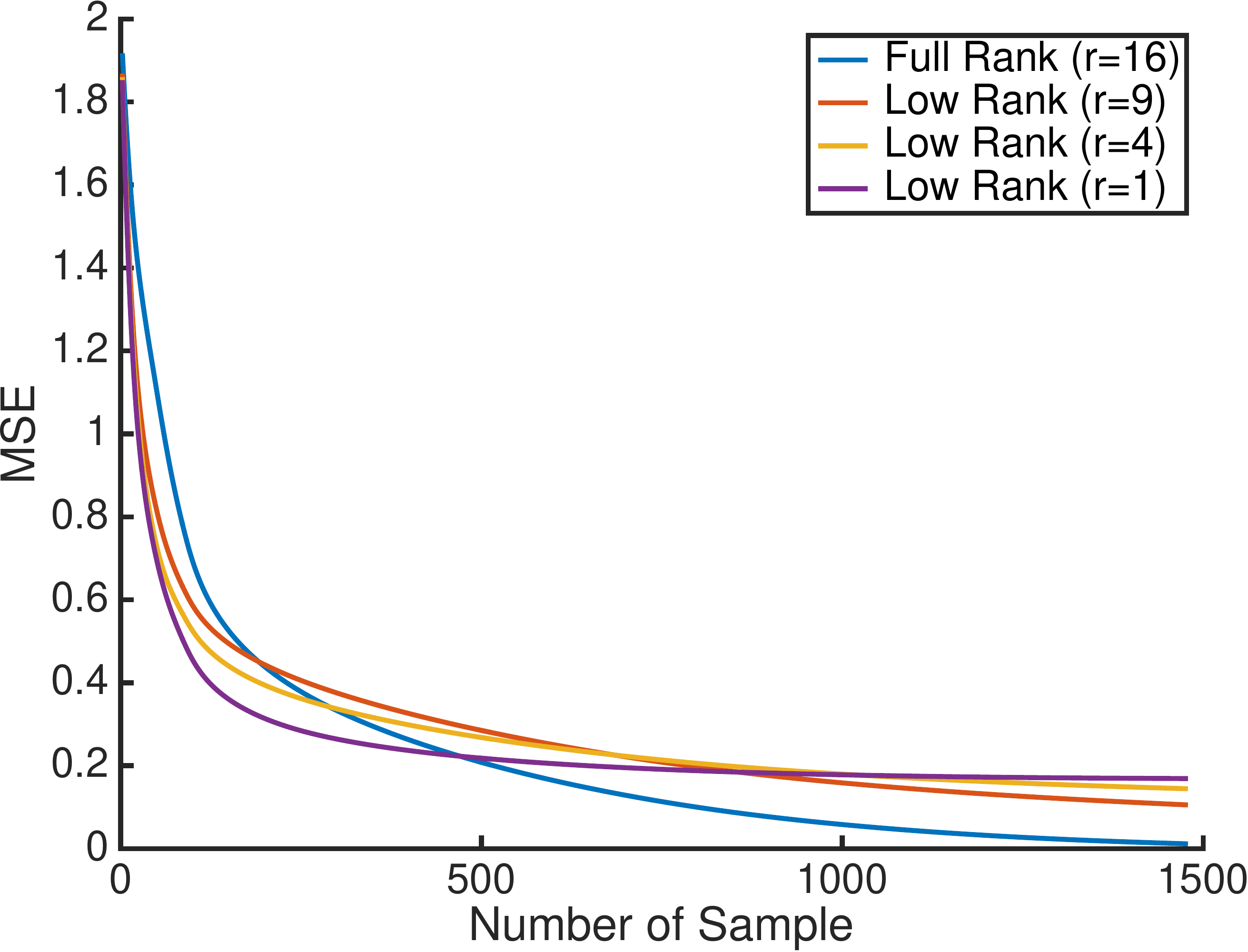}
			\label{fig:theory-low}
		}
		\subfigure[Low-rank $3$-mode MLGP] {
			\includegraphics[scale = 0.20]{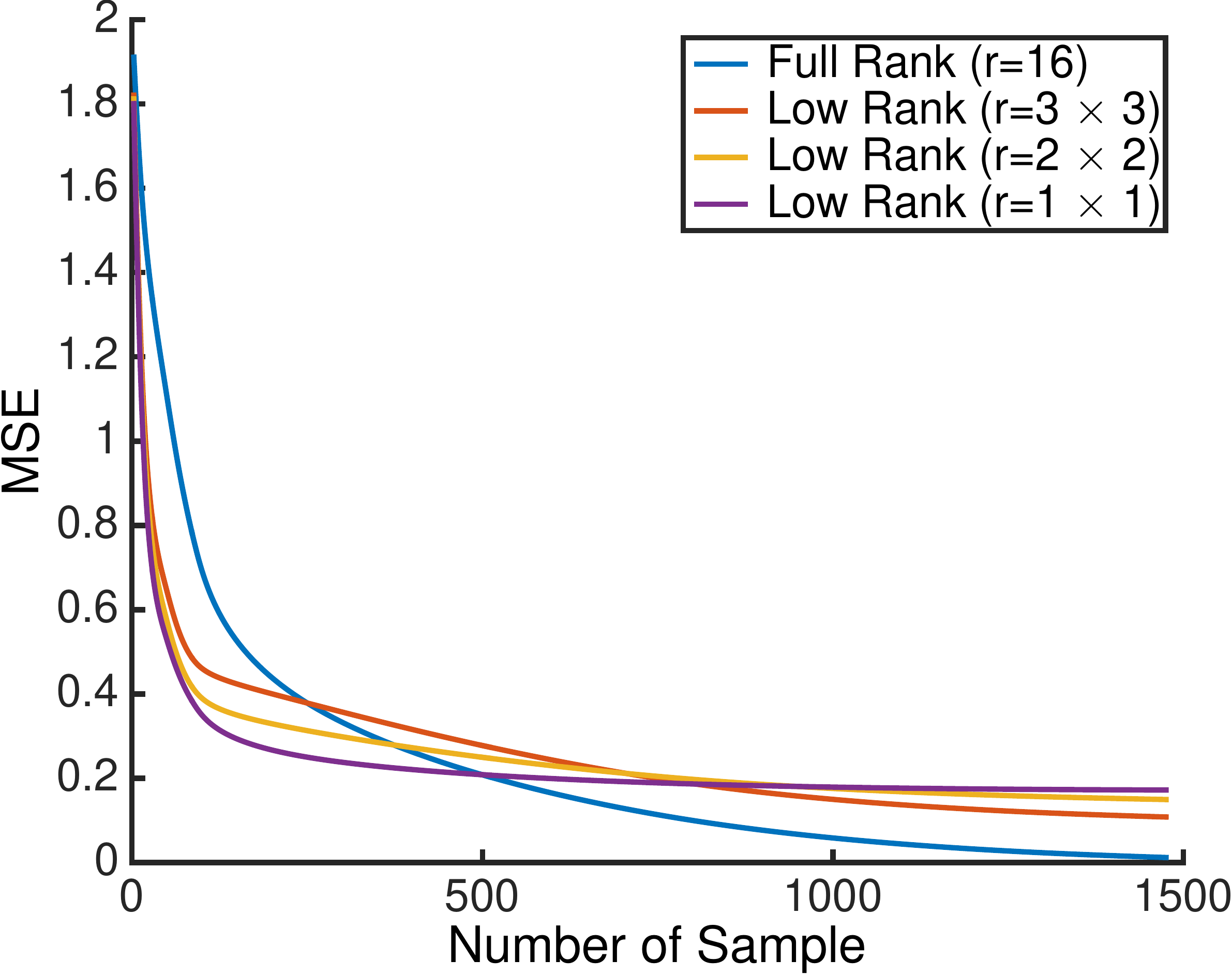}
			\label{fig:theory-kronecker}
		}
	\end{center}
	\caption{\subref{fig:theory-full} Theoretical and numerically simulated learning curve for task correlation $\rho=0.25,0.25,0.75$. \subref{fig:theory-low}  Learning curve for $2$-mode MLGP with low-rank approximation $r=9,4,1$. \ref{fig:theory-kronecker}  Learning curve for $3$-mode MLGP with low-rank approximation $r=9,4,1$.}
	\label{fig:exp-mlmtl}
\end{figure*}
We  study the theoretical properties of MLGP, which also shed light on the properties of existing tensor regression frameworks.

We first bound the excess risk of MLGP and derive the oracle inequality. Consider a  tensor of functionals $\T{W}$  and define a space $\T{C}_N$ with sample size $N$:
\begin{eqnarray*}
 \T{C}_N = \{ \T{W}: \T{W} = \T{S}\times_1 \M{U}_1 \times_2 \M{U}_2 \times_3 \M{U}_3,  \\ \|\T{S}_{(1)}\|_\star = \smallO\Big(\frac{N}{T_2 T_3+\log (T_1 T_2 T_3) }\Big)^{1/4} \} 
\end{eqnarray*} 
where $\|\cdot\|_\star$ denotes the matrix nuclear norm. The following proposition states the oracle inequality:
\begin{proposition}
	\label{thm:proposition}
Let $\hat{\T{W}}$	be the estimator that minimizes the empirical risk $ \hat{\loss}( f(\T{X}, \W); \T{Y}) $ over the space of  functional tensors $\T{W} \in  \T{C}_N$, then the excess risk, defined as $\T{L}$ satisfies:
\begin{eqnarray*}\T{L}(\hat{\T{W})} - \text{inf}_{\T{W} \in \T{C}_N}(\T{L}(\T{W})) \xrightarrow{P} 0\end{eqnarray*}

\end{proposition}
 \proof   Denote $\E[\text{cov}(\T{Y},   \M{U}_1(\T{X})] =  \M{\Sigma}(\M{U}_1)$, we first bound the difference:
 \begin{eqnarray*}
 \loss(\W) - \hat{\loss}(\W) \leq C \max \{ 2,  \|  \T{S}_{(1)}  \|_\star^2\}  \|\M{\Sigma}(\M{U}_1) -   \hat{\M{\Sigma}}(\M{U}_1) \|_2 
 \end{eqnarray*}
 The empirical risk is:
 \begin{eqnarray*}
 \loss (\hat{\W})  - \loss (\W^\star)
 &\leq& [ \loss(\hat{\W}) - \hat{\loss}(\hat{\W})]   -  [ \loss (\W^\star) - \hat{\loss}(\W^\star)  ]\\ 
 &\leq&  \mathcal{O}\Big(    \|\T{S}_{(1)} \| _\star^2 \|\M{\Sigma}(\M{U}_1) -   \hat{\M{\Sigma}}(\M{U}_1) \|_2     \Big)
 \end{eqnarray*} 
if we assume $\|\T{S}_{(1)}\|_\star^2 = \smallO( \Big(\frac{N}{T_2 T_3 + \log (T_1T_2 T_3)} \Big)^{1/4})$, then $ \loss (\hat{\W})  - \loss (\W^\star) \leq \smallO(1)$. Details of the derivation are deferred to Appendix \ref{sec:derive-prop}. \qedsymbol

 
This shows the estimation error tending to zero under a scaling assumption on the sample size $N$ and the dimensions $\{T_m\}$. However, asymptotic results can only capture the large $N$ regime  and will not  apply for finite sample sizes  in practice.  The following theorem states the explicit form of the non-asymptotic learning curve for the MLGP model under full-rank and low-rank scenarios: 
\begin{theorem}
	\label{thm:theorem}
 Assume the eigenfunction decomposition for the data-dependent part of covariance $\phi(\V{x})\M{K}_1 \phi(\V{x'})^\top = \sum_i \lambda_i \psi_i(\V{x})\psi_i(\V{x'})^\top$, denote $\M{\Lambda}$ as the diagonal matrix of $\{\delta_{i,j}\lambda_i \}$, the average case learning curve for  MLGP of single task $t$ satisfies
	\[ \epsilon(N)_t = \tr  \ \M{P}_{t_1,\cdots,t_M} \Big(  \M{\Lambda'}^{-1} + \sum_{s=1}^T \text{diag}(\frac{n_{s}}{\sigma^2_{s} + \epsilon_{s}})\M{P}_{s_m}\Big)^{-1}\]
when $\M{\Lambda'}$ is full-rank
{\small
		\[ \epsilon(N)_t = \tr\ \M{P}_{t_1,\cdots,t_M}\Big( \M{\Lambda'}-\Big(\sum_{s=1}^T \text{diag}(\frac{\sigma^2_{s} + \epsilon_{s}}{n_{s}})\M{P}_{s_m} + \M{\Lambda'}\Big)^{-1} \M{\Lambda'}^2\Big) \]
 }
when $\M{\Lambda'}$ is rank-deficient,  where $\M{P}_{t_1,\cdots,t_M}$ is the linear operator that maps index $t$ to a set of indices $\{t_m \}$, and $\M{\Lambda'} = \otimes_{m=2}^M \M{K}_m \otimes \M{\Lambda}$.
	\end{theorem}

\proof
The Bayes error, defined as $\hat{\epsilon} =	\E_{\V{x}}[(\V{w} -\hat{\V{w}})^2] $, has the following form  for the low-rank case:
	\begin{equation}
		\hat{\epsilon} =   \text{tr}\ (\M{\Lambda}) - \text{tr}(\M{D}+ \M{\Psi}\M{\Lambda}\M{\Psi}^\top)^{-1}\M{\Psi}\M{\Lambda}^2\M{\Psi}^\top  
        \label{eqn:be-single-full}
	\end{equation}
and    
	\begin{equation}
		\hat{\epsilon} = \text{tr}\ (\M{\Lambda}^{-1} + \M{\Psi}^\top \M{D}^{-1}\M{\Psi})^{-1}  \label{eqn:be-single-low}
	\end{equation}
for the full-rank case.  And $\M{\Lambda}$ and $\M{\Psi}$ are the eigen-components of the covariance.  The size of $\M{\Lambda}$ is equal to the number of kernel eigenfunctions. When the GP has a non-degenerate kernel, $\M{\Lambda}$ is full-rank. We can apply the Woodbury lemma to Equation  \ref{eqn:be-single-full}, which yields a  simplified version as  in Equation \ref{eqn:be-single-low}.  

Using method of characteristics \citep{sollich2002learning}, we can obtain a corresponding lower bound for the average case learning curve: 
\begin{equation}
	\epsilon(N)  = \text{tr}(\M{\Lambda}) - \text{tr} (\frac{\sigma^2+ \epsilon}{N}  \M{I} +\M{\Lambda})^{-1} \M{\Lambda}^2 \label{eqn:lc-single-full}
	\end{equation}

	\begin{equation}
	\epsilon(N)  = \text{tr}(\M{\Lambda}^{-1}+  \frac{N}{\sigma^2 + \epsilon} \M{I})^{-1}  \label{eqn:lc-single-low}
	\end{equation}

For MLGP, due to the task hierarchy, a task index  $t$  is projected to a set of indexes $\{t_m\}$ along different modes of a tensor. Define the projection on $m$th mode as  $\M{P}_{t_m} = \V{e}_{t_m} \V{e}_{t_m}^\top$, where $\V{e}_{t_m}$ is a unit vector with all zero but $t_m$ th entry as one.  Assume  eigenfunction decomposition for the data-dependent part of covariance $\phi(\V{x})\M{K}_1 \phi(\V{x})^\top = \sum_i \lambda_i \psi_i(\V{x})\psi_i(\V{x})^\top$, we have 
\begin{eqnarray*}
 \M{K}_{jk} = \prod_{m=2}^M\M{K}_{m,(\tau_j,\tau_k)}  \sum_i \lambda_i  \delta_{\tau_j, t}\psi_i(\V{x}_j) \delta_{\tau_k, t}\psi_i(\V{x}_k)^\top\\
 \M{K}= \M{\Psi} (\otimes_{m=2}^M \M{K}_m \otimes \M{\Lambda})\M{\Psi}^\top =  \M{\Psi} \M{\Lambda}'\M{\Psi}^\top
\end{eqnarray*}
where $\tau_j$ is the task index for $j$ th example, further projected to the mode-wise indexes. Augmented eigenfunction matrix $\M{\Psi}_{j,it} = \delta_{\tau_j, t}\psi_i(\V{x}_j)$ accounts for missing data, where the column index of $\M{\Psi}$ runs over all eigenfunctions and all tasks. For task $t$, denote $k_t (\V{x},\cdot) = k(\V{x}_t,\cdot)$
  \[\E_{\V{x}}[k_t(\V{x}, \M{X}) k(\V{X}, \M{x}_t)] = \M{\Psi}(\otimes_{m=2}^M(\M{K}_{m}\M{P}_{t_m} \M{K}_{m})\otimes \M{\Lambda}^2)\M{\Psi}^\top\]
where $\M{P}_{t_m}$ is the $m$ th mode index for task $t$.
The Bayes error can be written as:
\begin{eqnarray*}
	\label{eqn:be-multi-full}
 \hat{\epsilon}_t =& \E_{\V{x}}[k_t(\V{x},\V{x} )]  -  \E_{\V{x}}[ k_t(\V{x},\M{X} ) (\M{K}+\M{D})^{-1} k_t(\M{X},\V{x} ) ] \nonumber 
 \end{eqnarray*}
 For the first term
 \begin{eqnarray*}
\E_{\V{x}}[k_t(\V{x},\V{x} )] &=& \prod_{m=2} \V{e}_{t_m}^\top\M{K}_{m}\V{e}_{t_m}\E_{\V{x}}[\phi(\V{x})\M{K}_1\phi(\V{x})^\top]\\
 &=&\text{tr}\  \otimes_{m=2}^M\M{P}_{t_m} \M{K}_{m}\otimes\M{\Lambda}
\end{eqnarray*}
For the second term
\begin{align*}
&\E_{\V{x}}[ k_t(\V{x},\M{X} ) (\M{K}+\M{D})^{-1} k_t(\M{X},\V{x} ) ] = \\
\text{tr}\ (\M{D}+\M{\Psi} &\M{\Lambda}'\M{\Psi}^\top)^{-1}\M{\Psi}(\otimes_{m=2}^M(\M{K}_{m}\M{P}_{t_m} \M{K}_{m})\otimes \M{\Lambda}^2)\M{\Psi}^\top  
\end{align*}
 With $\otimes_m \M{P}_{t_m} = \M{P}_{t_1,\cdots, t_M}$, compare Equation \ref{eqn:be-multi-full} with Equation \ref{eqn:be-single-full}, we have
 \begin{eqnarray*}
&\hat{\epsilon}_t = \M{P}_{t_1,\cdots, t_M} \Big( \text{tr}(\M{\Lambda'}) - \text{tr}(\M{D}+ \M{\Psi}\M{\Lambda'}\M{\Psi}^\top)^{-1}\M{\Psi}\M{\Lambda'}^2\M{\Psi}^\top  \Big)\\
& \M{\Lambda}' = \otimes_{m=2}^M \M{K}_m \otimes \M{\Lambda} 
 \end{eqnarray*} 
The Bayes error of task $t$  is that of all tasks projected to each of its mode-wise task indices. Using an  analogous method of characteristic curves, we can obtain a set of self-consistency equations for the learning curve of MLGP (see Appendix \ref{sec:derive-thm} for details). \qedsymbol

Theorem \ref{thm:theorem} indicates the performance dependency of MLGP, hence tensor regression, on the eigenvalues of the covariance function as well as the task correlation matrix. When the number of examples for all tasks becomes large, the Bayes errors  $\hat{\epsilon}_t$ will be small and eventually be negligible compared to the noise variances  $\sigma_t$.  This also reflects a commonly accepted claim for the asymptotic useless of multi-task learning: when the number of samples becomes large, the learning curves would come close to single task learning, except for the fully corrected case.

We further conduct numerical simulations  to better understand the derived   learning curve.  Consider  the case with 16 identical tasks, and  set the task correlation matrix $\otimes_{m=2}^M\M{K}_m$  to have $\rho$ everywhere  except for the principal diagonal elements.
Assuming all the  tasks are identical,  Figure \ref{fig:theory-full} compares the theoretic  learning curve with the numerically simulated learning curve for different task relatedness. The theoretical learning curves generally lay slightly below the actual  learning curves, providing a  tight lower bound. With a higher value of $\rho$,  tasks share higher interdependence, resulting in faster convergence  w.r.t. Bayes error.

Figure \ref{fig:theory-low} shows the learning curve for $2$-modes MLGP with different low-rank  approximations  with $R_m=[1,4, 9,16]$. The low-rankness alleviates the noise variance error, leading to a faster convergence rate but  eventually converges to a solution with a larger approximation gap.  Figure \ref{fig:theory-kronecker} displays the learning curves for the $3$-modes MLGP model, with the similar low-rank approximation. We observe that under the same rank assumption, the $3$-mode MLGP imposes a stronger prior, leading to  superior performances over $2$-model MLGP with sparse observations.

\subsection{Relation to  Other Methods}
It turns out that for multi-output regression, where all the tasks share the  same inputs $\M{X}_0 \in \R^{n_0 \times D}$, we can write  $\M{X} = \M{X}_0 \otimes \M{I}_T $, and noise becomes $\M{D}= \text{diag}([\sigma_1,\cdots, \sigma_T]) \otimes \M{I}_{n_0} $.  The covariance  $\M{K} = (\otimes_{m=2}^M \M{K}_m) \otimes \phi(\M{X}_0 )\M{K}_1\phi(\M{X}_0)^\top  =  (\otimes_{m=2}^M \M{K}_m) \otimes \M{K}_x$, where $\otimes_{m=2}^M \M{K}_m$ encodes task similarity and $\M{K}_x$  is the kernel matrix over inputs $\M{X}_0$.  When the number of  modes $M =2$, the model reduces to the multi-task Gaussian process (MTGP) model  with free-form parameters \citep{bonilla2007multi}. Here we factorize over Kronecker product operands as   the low-rank approximation  while  MTGP uses Nystr\"{o}m approximation.

The multi-linear kernel $\phi(\M{X})(\otimes_{m=1}^M \M{K}_m )\phi(\M{X})^\top$ allows us to compute $\{\M{K}_m \}_{m=1}^M$  separately, which avoids inversion of the big covariance matrix $\M{K}$. This property  has also been exploited in \citep{wilson2014fast} for multidimensional pattern extrapolation (GPatt). In there, inputs are assumed to be on a multidimensional grid $\V{x} \in \T{X} = \T{X}_1 \times \cdots \times \T{X}_M$,  the covariance matrix has decomposition $\M{K} = \otimes_{m=1}^M \M{K}_m$ where each factor $\M{K}_m $ is a kernel matrix over the space $\T{X}_m$. The difference is that  we use  Kronecker products to learn multi-directional task correlations while GPatt performs kernel learning for each dimension of the inputs.

\begin{figure}[h]
\begin{center}
\subfigure[Restaurant MSE]{
\includegraphics[width=0.45\columnwidth]{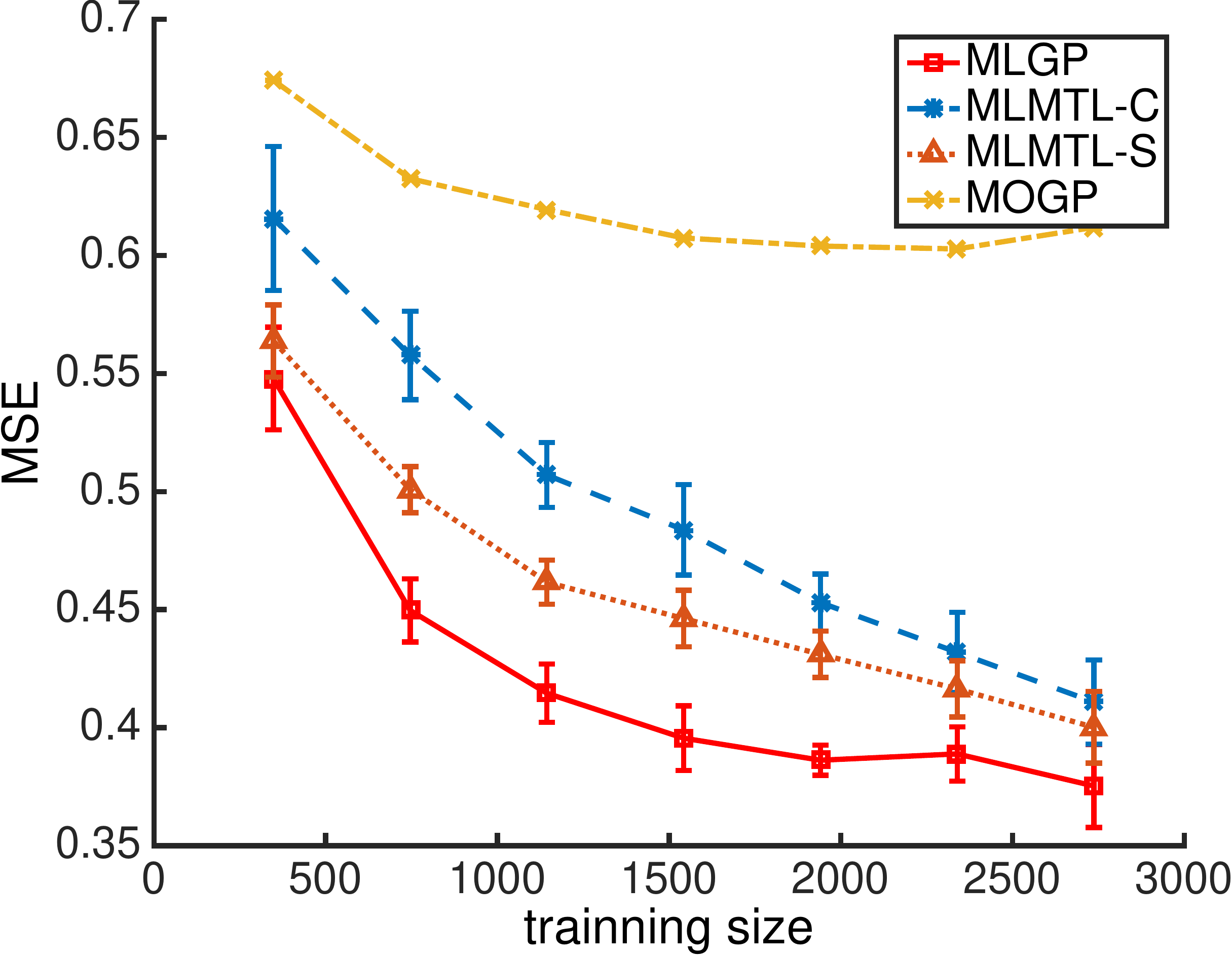}
\label{fig:restaurant}
}
\subfigure[School EV]{
	\includegraphics[width=0.45\columnwidth]{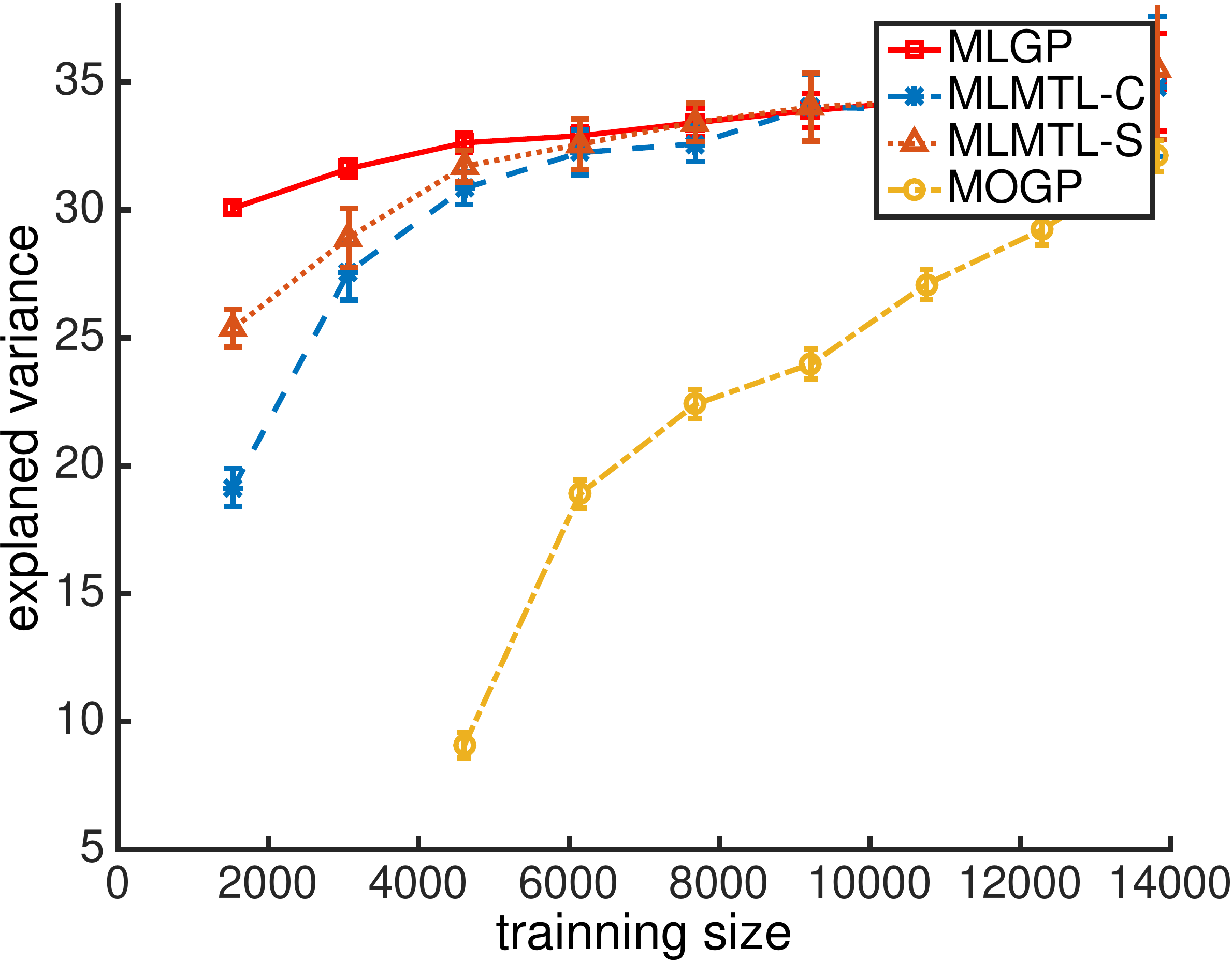}
\label{fig:school}
}
\end{center}
		\vspace{-5mm}
\caption{Multi-linear multi-task learning  benchmark comparison \subref{fig:restaurant} mean square error  on the restaurant dataset. \subref{fig:school}  expected variance on the school dataset.  w.r.t sample size  for MLGP and baselines. }
\label{fig:exp-mlmtl}
\end{figure}

\begin{figure*}[ht]
	\begin{center}
		\subfigure[\textsf{PRCP}]{
			\includegraphics[width = 0.17\textwidth]{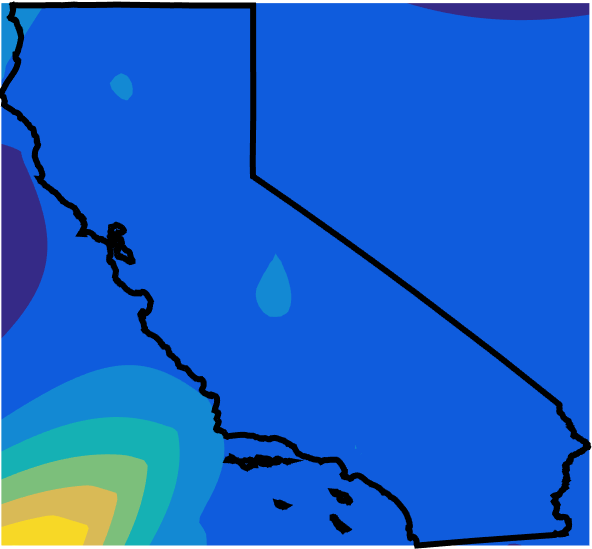}
			\label{fig:prcp}
		}
			\subfigure[\textsf{TMAX}]{
				\includegraphics[width = 0.17\textwidth]{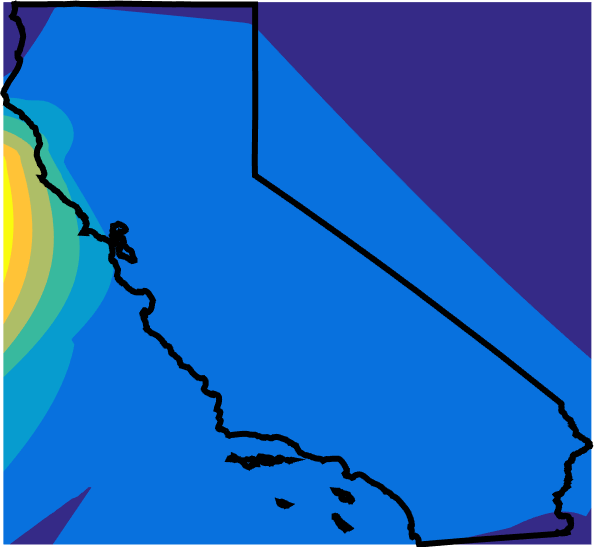}
				\label{fig:tmax}
			}
			\subfigure[\textsf{TMIN}]{
				\includegraphics[width = 0.17\textwidth]{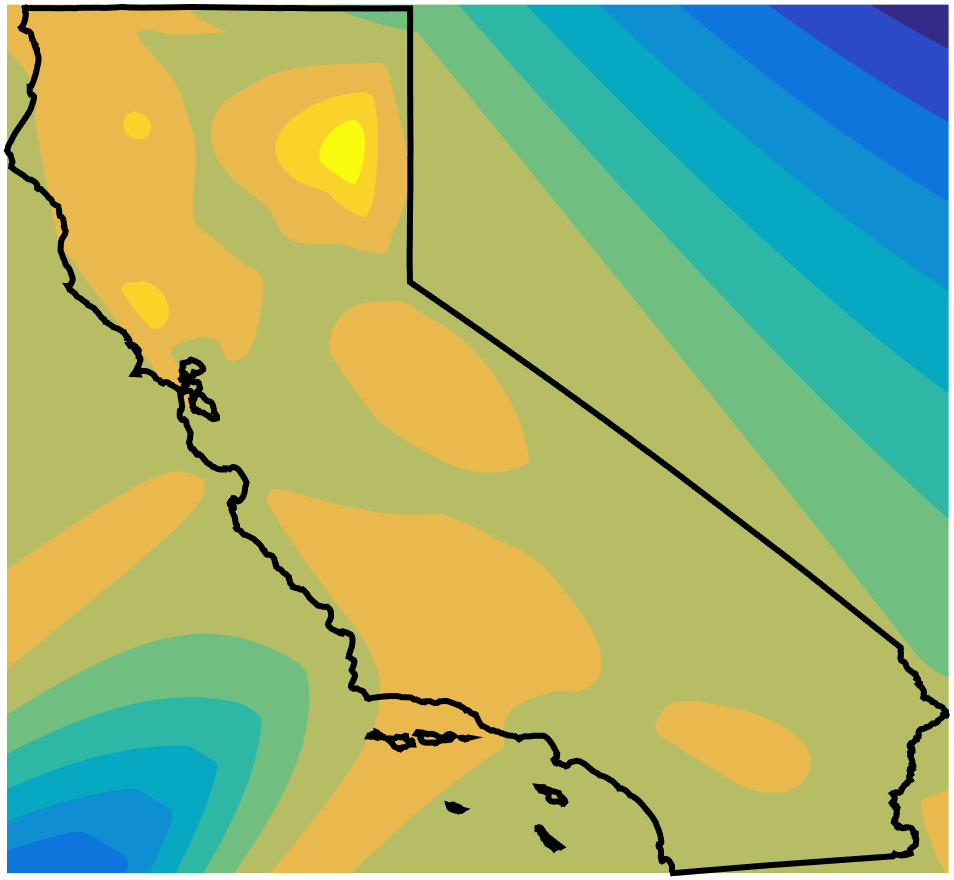}
				\label{fig:tmin}
			}
		\subfigure[\textsf{SNOW}]{
			\includegraphics[width = 0.17\textwidth]{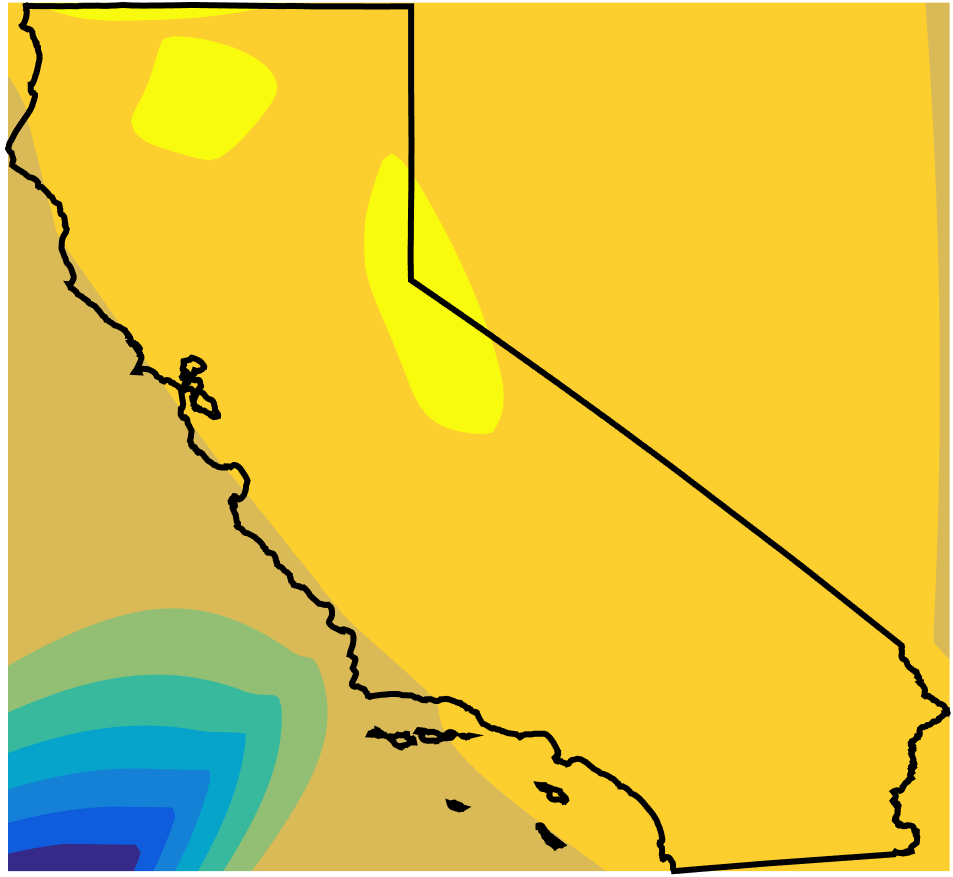}
			\label{fig:snow}
		}
		\subfigure[\textsf{SNWD}]{
			\includegraphics[width = 0.17\textwidth]{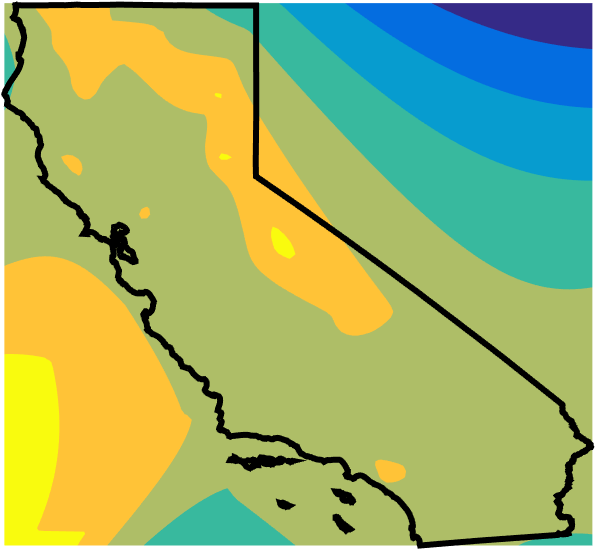}
			\label{fig:snwd}
		}
	\end{center}
	\caption{Contour plots  for the MLGP predictive variance w.r.t precipitation \textsf{PRCP}, max temperate \textsf{TMAX} min temperate \textsf{TMIN}, snowfall \textsf{SNOW}and snow depth \textsf{SNWD}.  Yellow is high variance  and blue means low variance.}
	\label{fig:exp-contour}
\end{figure*}

	\section{Experiments}

We conduct  experiments for a series  of tensor regression applications and demonstrate comparable prediction performances of MLGP with confidence intervals.


\subsection{Multi-linear Multi-task learning}
 We  evaluate on  two benchmark datasets  for MLMTL:  school exam scores  and restaurant ratings. School exam scores contain $15,362$ students exam records with $21$ features from $139$ schools across $3$ years.  Each task is defined as the prediction of the exam score of a student from  a specific school in one year given school-student attributes. Restaurant ratings contain $3,483$ rating records with $45$ features from $138$ consumers for $3$ aspects. A task is defined as prediction of rating for an aspect from a specific consumer given restaurant attributes.

We compare with the following baselines.
(1) \textit{MLMTL-C} \citep{romera2013multilinear}: latent trace norm optimization with alternating direction method of multipliers (ADMM)
(2)\textit{MLMTL-S} \citep{wimalawarne2014multitask}: scaled latent trace norm optimization with ADMM, 
and (3) \textit{MOGP} \citep{alvarez2011computationally}: multi-output Gaussian process with DTC variational kernel.   As all methods consider linear regression tasks, we use linear kernel  MLGP  as a fair comparison. For MOGP, we use $20$ inducing points.


We randomly selected from a range of $10\%$ to $80\%$ of the entire data set as the training set. We selected $10\%$ instances as the validation set and the rest was used as the test set. The regularization parameter for each norm was selected by minimizing the mean squared error on the validation set.  We repeat the experiments for $10$ times and average the results. All the baselines are the implementations of the original  authors. 

Figure \ref{fig:restaurant} shows the  restaurant rating prediction mean square error (MSE) for different methods over number of training samples. Figure \ref{fig:school} demonstrates the expected variance (EV) for the task of school exam score prediction. We observe superior performances of MLGP on restaurant data and comparable results for school data. In parti=cular, when the size of the training data is small, MLGP shows significant advantages for both tasks. This justifies the benefit of MLGP for sparse observations. 
%
%

\begin{table*}[ht]
	\caption{Mean square error comparison of MLGP and baselines for  spatio-temporal forecasting on $4$ datasets with $10\%$ testing set. Tensor regression models use VAR-3 with moving window.}
	\label{tb:exp-forecasting}
	\centering
	\begin{tabular}{l c c  c  c  c c  }
		\toprule
		Dataset    &MLGP  &  Greedy  & MLMTL-C & MLMTL-S   &  MTL-Trace \\
		\midrule
		USHNC-US  &$0.8973 \mp{0.0008}$ &  $0.9069$&$0.9528$ & $0.9543$ & $0.9273$ \\
		CCDS     &$0.8498 \mp{0.0013}$ &$0.8325$ &$0.9105$ & $0.8394$ & $0.8632$ \\
		FSQ      &$0.1248 \mp{0.0006}$ & $0.1223$ &$0.1495$ & $0.1243$ & $0.1245$ \\
		YELP   &$1.0725 \mp{0.0007}$ & NA &$1.0857$ & $1.0876$ & $1.0736$ \\
		\bottomrule
	\end{tabular}
\end{table*}

\subsection{Spatio-temporal Forecasting}
Spatio-temporal forecasting has been shown to be a special case of tensor regression, with an additional spatial Laplacian matrix \citep{yu2014fast}. We evaluate the spatio-temporal forecasting performance  for  $4$ datasets reported in the original paper. For all the datasets, each variable is  normalized by removing mean and dividing by variance.  A third-order vector auto-regressive (VAR-3) model is employed for multi-variate time series modeling.  We perform an $80/20$ split along the time direction for training/testing and use validation to select the rank hyper-parameter.

Table \ref{tb:exp-forecasting} displays the forecasting MSE comparison. We compare with the reported best algorithm Greedy \citep{yu2014fast}  for this task.  We also include matrix multi-task learning with trace-norm regularization (MTL-Trace) to justify the benefit of the tensor-based approach. For all the $4$ datasets,  MLGP obtains similar prediction accuracy as Greedy. The predictive variance from MTGP directly provides empirical confidence intervals, which we append to the MSE.

To better understand  the learned predictive distribution, we use  a fine-grained USHCN dataset from California \citep{yu2016learning} and visualize the predictive variance  of different locations on the map.  We interpolate the variance values  across  locations and draw the contour plots. Figure \ref{fig:exp-contour} shows the contour plot for $54$ locations of $5$ climate variables.   We observe interesting correlations between the predictive variance and geographical attributes. For example, precipitation (PRCP) and maximum temperate (TMAX) have relatively low-variance due to the subtropical climate in California. Snow depth (SNWD) demonstrates high variance along the mountains in Sierra Nevada.

\begin{figure}[t]
	\begin{center}
		\subfigure[XAU]{
			\includegraphics[width = 0.29\columnwidth]{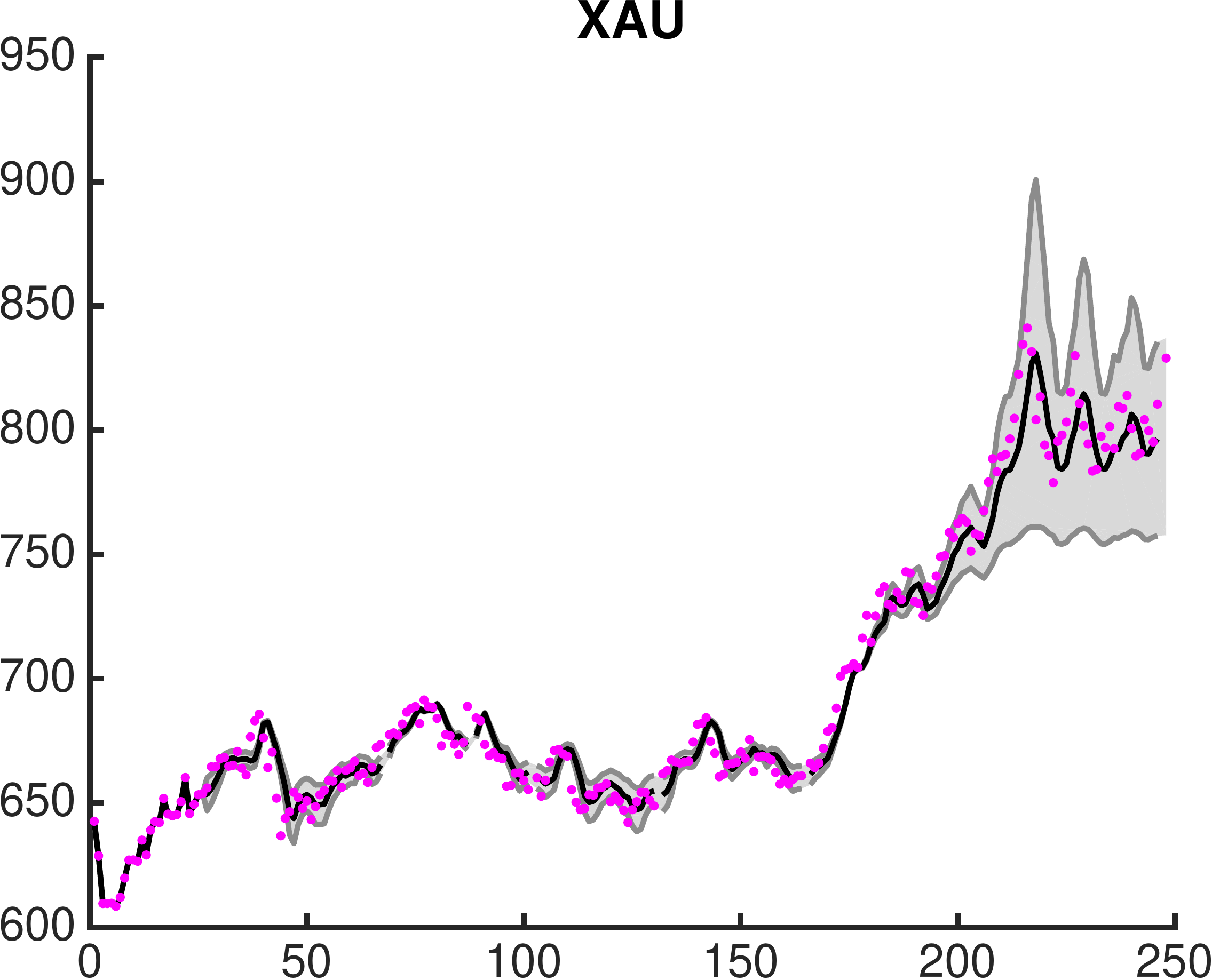}
			\label{fig:exp-fx_xau}
		}
		\subfigure[XAG]{
			\includegraphics[width = 0.29\columnwidth]{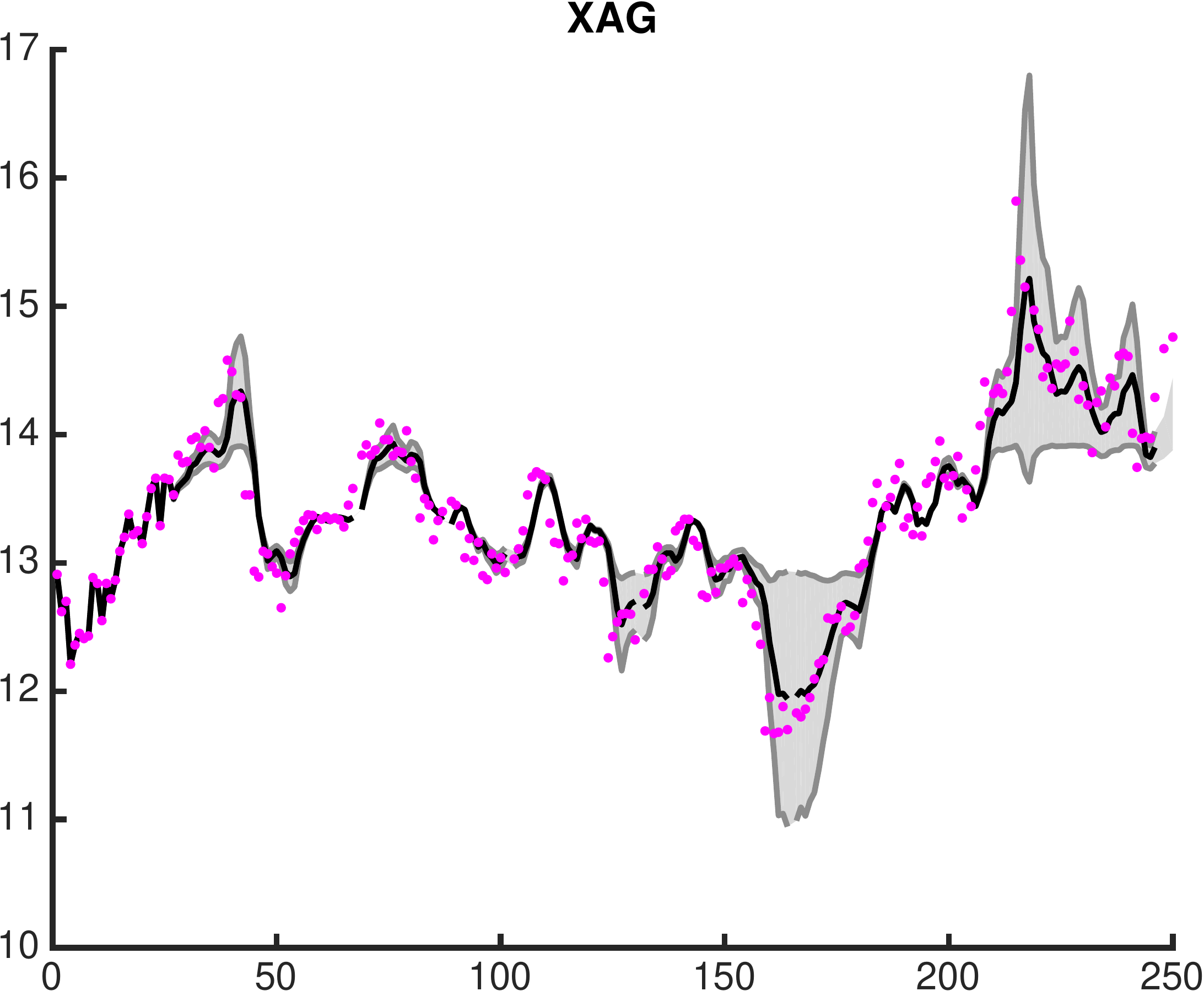}
			\label{fig:exp-fx_xag}
		}
		\subfigure[EUR]{
			\includegraphics[width = 0.29\columnwidth]{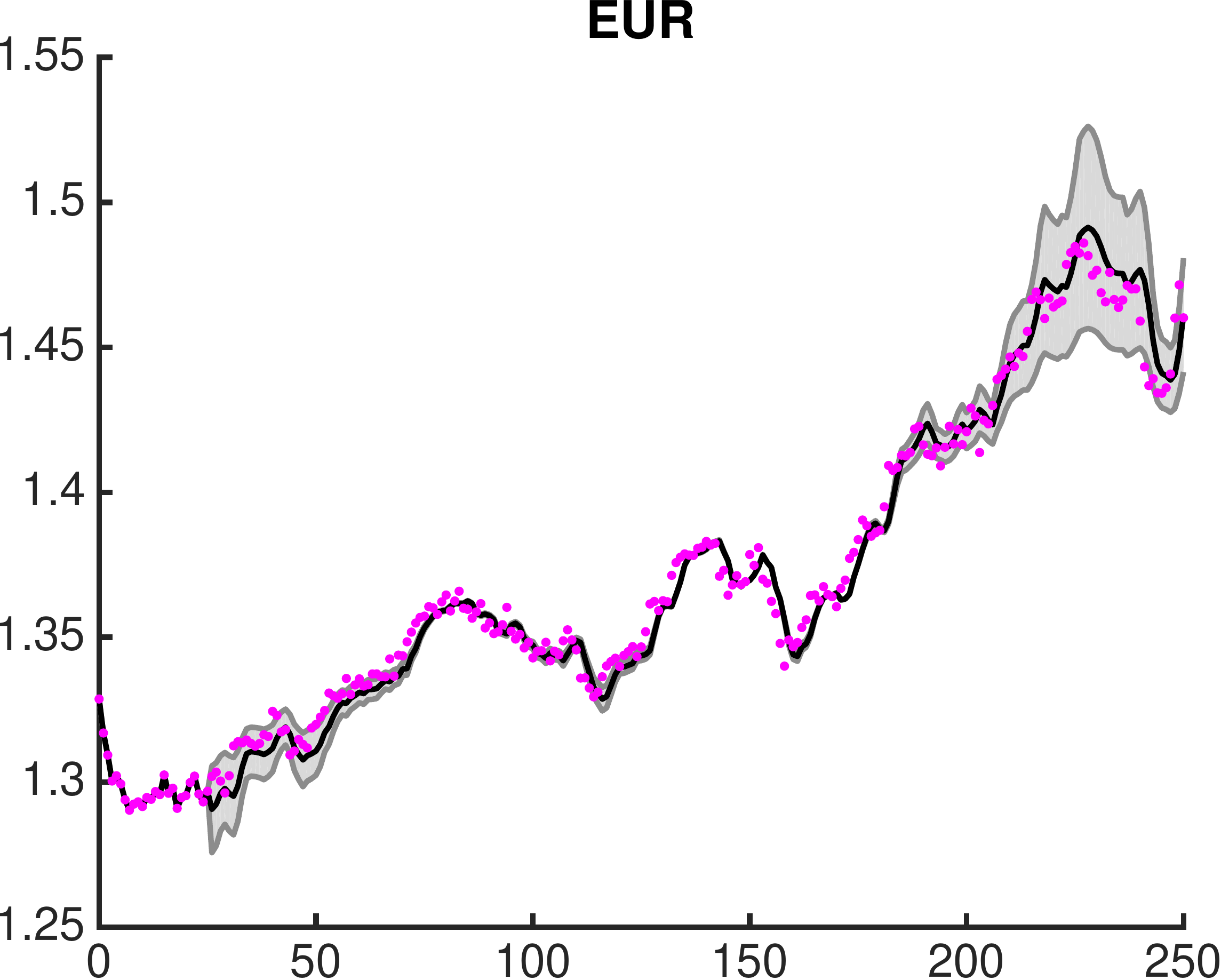}
			\label{fig:exp-fx_eur}
		}
		\subfigure[GBP]{
			\includegraphics[width = 0.3\textwidth]{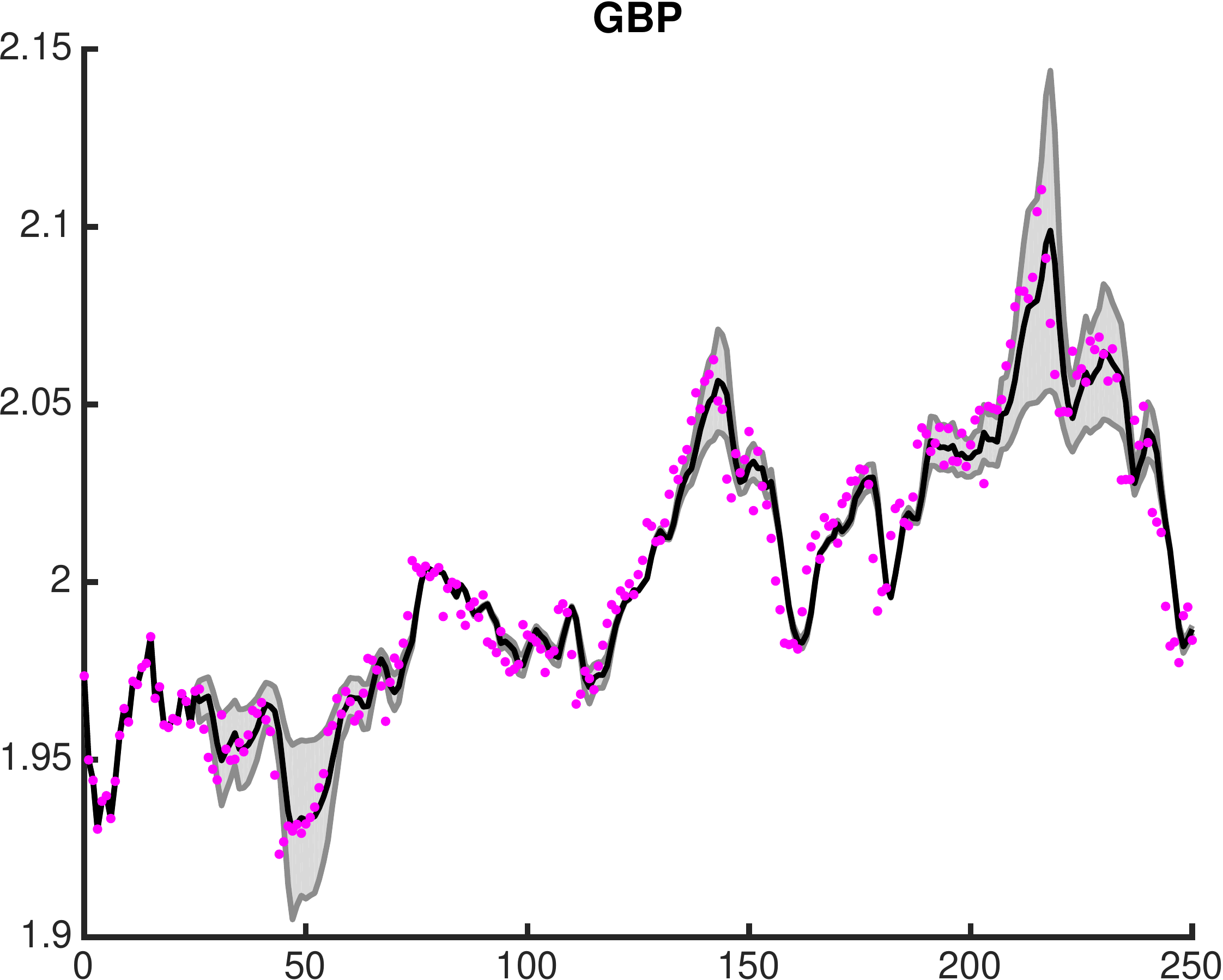}
			\label{fig:exp-fx_gbp}
		}
 		\subfigure[JPY]{
 			\includegraphics[width = 0.3\textwidth]{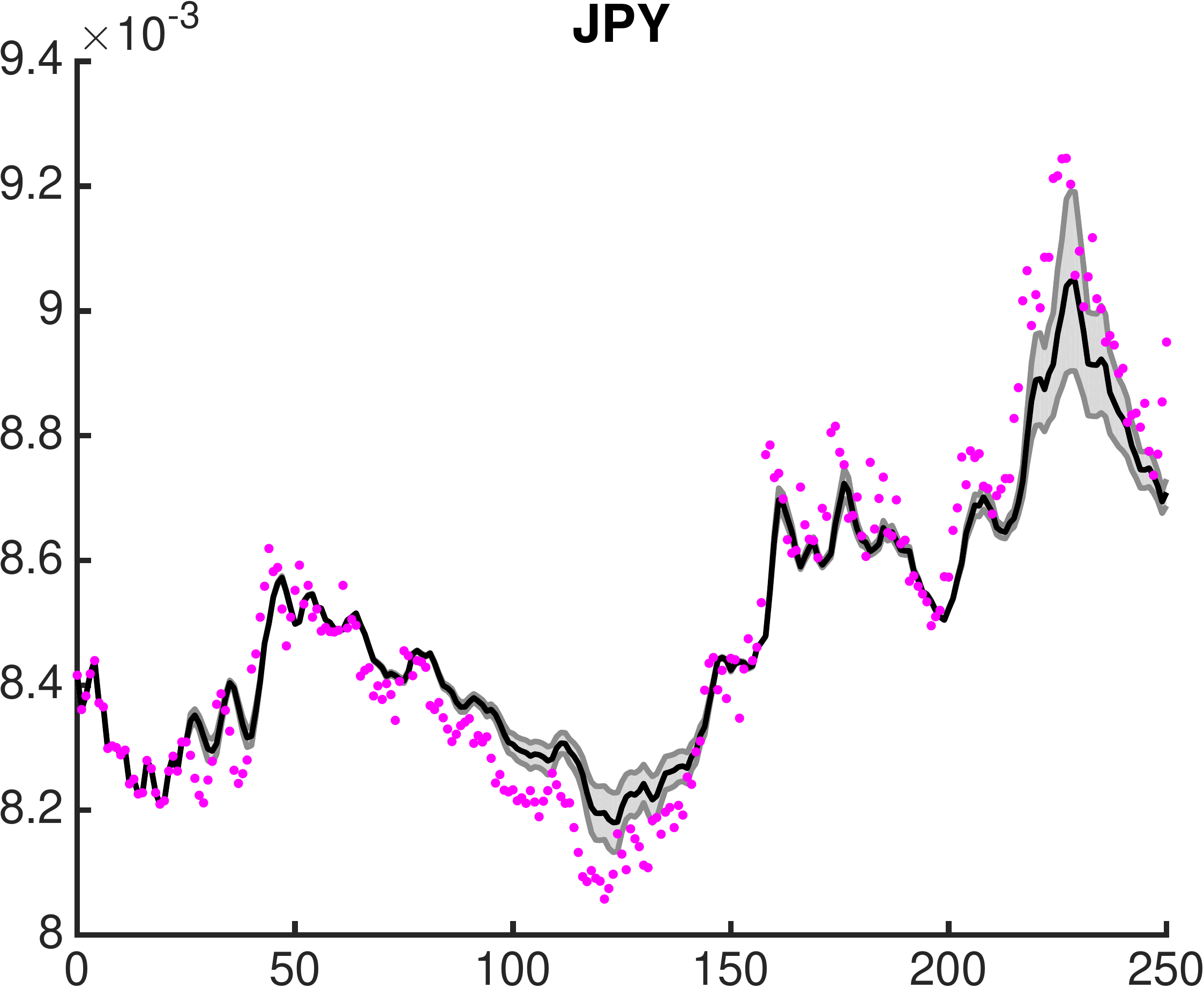}
 			\label{fig:exp-fx_jpy}
 		}
		\subfigure[HKD]{
			\includegraphics[width = 0.3\textwidth]{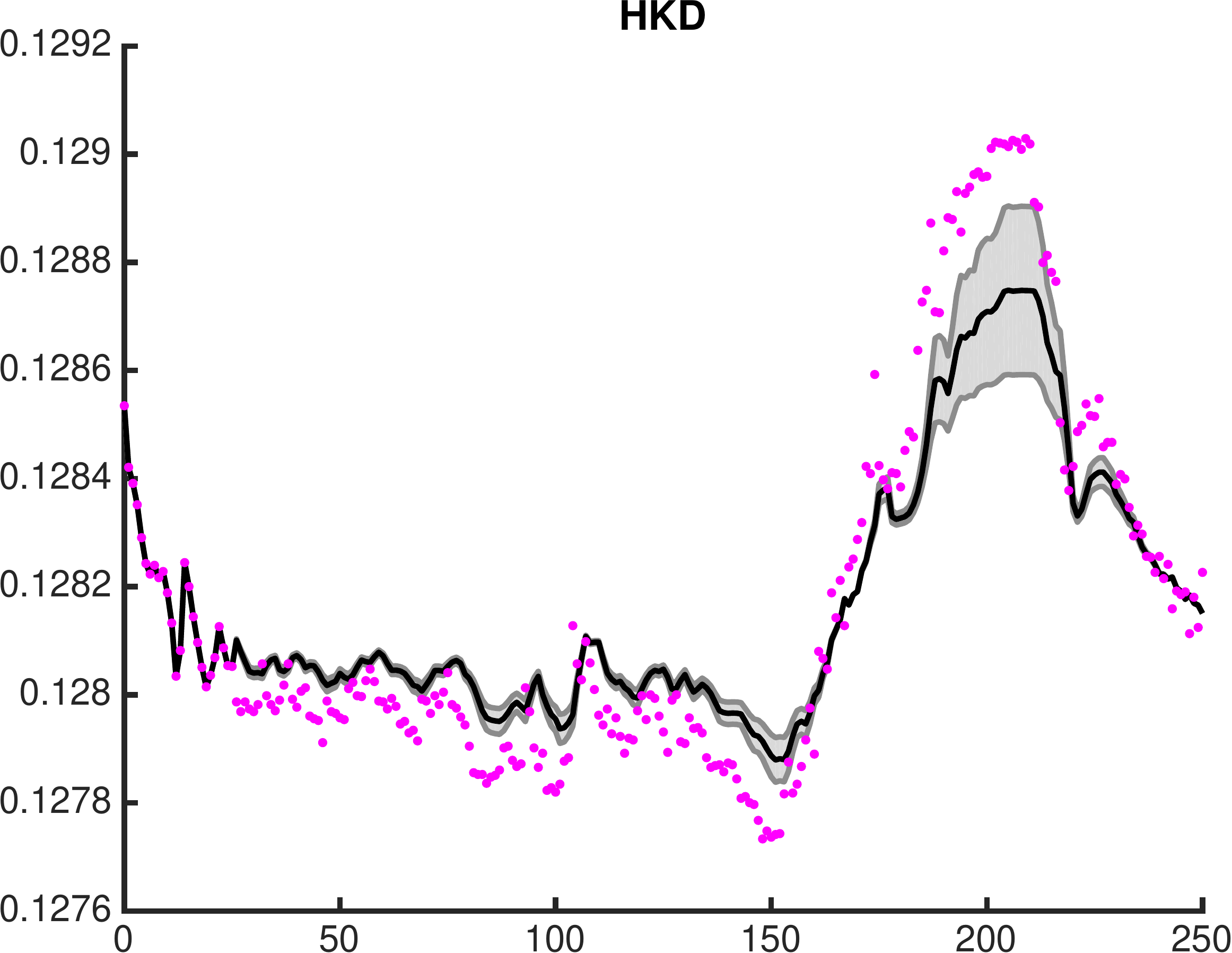}
			\label{fig:exp-fx_hkd}
		}
	\end{center}
	\caption{Predictive mean (solid line)  and variance (shaded area)  for foreign exchange rate of $6$  international currencies: XAU and EUR using from MLGP for $50$ time step ahead forecasting. Magenta points are observations.  }
	\label{fig:exp-timeseries}
\end{figure}


%

\subsection{Multi-output regression}
Multiple output regression concerns with the case when predictor tensor is shared among all of the responses. One  such application is the foreign exchange rate prediction task \citep{alvarez2011computationally}. The original dataset contains $3$ precious metals and $12$ international  currencies. To show the benefit of exploiting multi-directional task interdependence, we select the foreign exchange rate  of  $6$ international currencies ( EUR, GBP, CHF, JPY, HKD, KRW) and $3$ precious metals (gold, silver, and platinum), which forms three groups: precious metal, European currency and Asian currency.   The dataset consists of all the data available for the $251$ working days in the year of $2007$. 


We use the VAR-3 model for  all the low-rank tensor regression baselines.  MLGP achieves $0.0563$ MSE while best performance of low-rank tensor regression is $0.0657$. These results  are slightly worse than $0.0301$ of MOGP with PITC approximation. However, since MLGP does not require all the responses to be of equal size, it runs much faster than MOGP, which involves a missing value imputation step to satisfy the size constraint. To further interpret the learned model, we plot out the predictive mean and variance together with observations in Figure \ref{fig:exp-timeseries}. We observe high predictive variance  whenever the time series encounters sharp changes.

	\section{Discussion and Conclusion}	
	In this paper, we establish interesting connections between tensor regression and Gaussian processes. We develop a probabilistic counterpart: multi-linear Gaussian processes (MLGP). With the low-rank constraint, the  Bayesian estimator of MLGP  learns a smooth surrogate for the low-rank structure in tensor regression. Theoretical analysis shows its performance dependency on the eigenvalues of the covariance matrix  and task correlation.  Comparable (if not better) performance are observed in a series of real-world applications.  

This relationship hints upon our choice of tools for multi-way data analysis.  Tensor regression is fast and simple to implement. It is  guaranteed to output orthonormal basis of the latent subspaces but does not generate confidence intervals. MLGP, on the other hand, can better handle sparse observations, and is more versatile with kernels. In terms of future directions, one interesting question is to study the robustness of both methods under adversarial corruptions. 
One possible future direction is  to kernelize tensor regression, so it can go beyond the current linear model and share the same flexibility as Gaussian processes.  The other  interesting question is to study the robustness of both methods under adversarial corruption.  This would help understand how many corruptions the model can tolerate with arbitrary, and possibly severe or correlated errors in the covariance matrix. 


\balance
\bibliography{icml2017}
\bibliographystyle{plainnat}
	\newpage
	\appendix

\onecolumn{
	\section{ Supplementary: On the Equivalence of Tensor Regression and Gaussian Process}
	\label{sec:app-derive}
\subsection{Eigenvalue problem}
\label{sec:derive_eigen}
Let $\M{K} = \tilde{\M{U}}  \tilde{\M{U}}^\top$, take derivative over $\tilde{\M{U}}$, we obtain the stationary point condition:  
$\V{y}\V{y}^\top (\M{K}+ \M{D})^{-1}   \tilde{\M{U}}  = \tilde{\M{U}} $, 
Given the decomposition of $\tilde{\M{U}} = \M{U}_x \M{\Sigma}_x \M{V}_x^\top $, similar to \citep{lawrence2004gaussian}, we have
\begin{eqnarray*}
\V{y}\V{y}^\top (\M{K}+ \M{D})^{-1}  \tilde{\M{U}} &=&  \tilde{\M{U}} \\
\V{y}\V{y}^\top (\M{K}+ \M{D})^{-1} \M{U}_x \M{\Sigma}_x \M{V}^\top_x   &=& \M{U}_x \M{\Sigma}_x \M{V}^\top_x \\
\V{y}\V{y}^\top \M{U}_x ( \M{\Sigma}_x  + \M{D}\M{\Sigma}_x^{-1} )^{-1}  \M{V}^\top_x &=&  \M{U}_x \M{\Sigma}_x \M{V}^\top_x\\
\V{y}\V{y}^\top \M{U}_x &=&\M{U}_x (\M{\Sigma}_x^2  + \M{D} )
\end{eqnarray*}
which is a  eigenvalue problem  in the transformed space.

\subsection{Derivatives for the Optimization}
\label{sec:derive-opt}
Given that $\V{y} \sim N(\V{0},\M{K}+\M{D})$, where $\M{K} = \phi(\M{X} )\otimes_{m=1}^M \M{K}_m \phi(\M{X})^\top$. 

Decompose $\M{K}_m = \M{U}_m\M{U}_m^\top$, we have $\M{K}=\phi (\M{X})(\otimes_{m=1}^M \M{U}_m) (\otimes_{m=1}^M \M{U}_m^\top) \phi (\M{X})^\top$. 

Let $\tilde{\M{U}} = \phi (\M{X})(\otimes_{m=1}^M \M{U}_m) $, we have $\M{K} = \tilde{\M{U}}  \tilde{\M{U}}^\top$

The negative log-likelihood 
\begin{align*}
	L = \frac{1}{2} \V{y}^\top (\tilde{\M{U}}\tilde{\M{U}}^\top + \M{D})^{-1}  \V{y} +\frac{1}{2} \log \det (\tilde{\M{U}}\tilde{\M{U}}^\top + \M{D} ) + const 
\end{align*}

Based on Woodbury lemma, $(\tilde{\M{U}}\tilde{\M{U}}^\top + \M{D})^{-1}  = \M{D}^{-1} - \M{D}^{-1} \tilde{\M{U}} (\M{D} + \tilde{\M{U}}^\top  \tilde{\M{U}} )^{-1} \tilde{\M{U}}^\top $ as well as  matrix determinant lemma $\det (\tilde{\M{U}}\tilde{\M{U}}^\top + \M{D})=\det( \M{I}+\tilde{\M{U}}^\top \M{D}^{-1}\tilde{\M{U}} )\det(\M{D}) = \det(\M{D}+\tilde{\M{U}}^\top\tilde{\M{U}})$

Denote $ \M{\Sigma} =  \M{D}  + \tilde{\M{U}}^\top \tilde{\M{U}}  $, let $\V{w}= \M{\Sigma}^{-1} \tilde{\M{U}}^\top\V{y}$. The objective function can be rewrite as 
\begin{align*}
		L = \frac{1}{2} \M{D}^{-1}\V{y}^\top \V{y} -\frac{1}{2} \M{D}^{-1}\V{y}^\top\tilde{\M{U}} \M{\Sigma}^{-1} \tilde{\M{U}}^\top \V{y} +\frac{1}{2} \log \det(\M{\Sigma})+ const 
\end{align*}
Take derivative over $\M{U}_{m(i,j)}$, we  have 
\begin{align*}
	\frac{\partial L}{\partial \M{U}_{m(i,j) }} = \text{tr}[(\frac{\partial L}{\partial \tilde{\M{U}}})^\top(\frac{\partial \tilde{\M{U}}}{\partial \M{U}_{m(i,j) }})],  
\quad 
	\frac{\partial L}{\partial \tilde{\M{U} }} 	= \tilde{\M{U}}(\M{\Sigma}^{-1} +  \V{w}\M{D}^{-1}\V{w}^\top  )^{-1} - \V{y} \M{D}^{-1}\V{w}^\top
	\end{align*}
\begin{align*}
	\frac{\partial \tilde{\M{U}}}{\partial \M{U}_{m(i,j) }}	 =  \frac{\partial \phi(\M{X})}{\partial \M{U}_{m(i,j)  }} 	(	\M{U}_M\otimes \cdots \frac{\partial \M{U}_m}{\partial \M{U}_{m(i,j) }}\cdots \otimes \M{U}_1) =  \frac{\partial \phi(\M{X})}{\partial \M{U}_{m(i,j)  }}  (	\M{U}_M\otimes \cdots \M{O}_{m (i,j)}\cdots \otimes \M{U}_1)
	\end{align*}
Here $ \M{O}_{m (i,j)} = \V{e}_i \V{e}_j^\top$ is a matrix with all zeros, but the $(i,j)$th entry as one. 

The predictive distribution:
   $p(y_\star | \V{x}_\star, \M{X}, \V{y}) \sim N (\V{\mu}_\star, \sigma_\star)$:
   \begin{align*}
   	\V{\mu_\star} &=  \V{k}(\V{x}_\star,\M{X} ) (\M{D}^{-1} - \M{D}^{-1} \tilde{\M{U}} (\M{D} + \tilde{\M{U}}^\top  \tilde{\M{U}} )^{-1} \tilde{\M{U}}^\top )\V{y}\\
   	\V{\sigma}_{\star}& =   \V{k}(\V{x}_\star,\V{x}_\star )-  \V{k}(\V{x}_\star,\M{X} )(\M{D}^{-1} - \M{D}^{-1} \tilde{\M{U}} (\M{D} + \tilde{\M{U}}^\top  \tilde{\M{U}} )^{-1} \tilde{\M{U}}^\top )\V{k}(\M{X},\V{x}_\star )
   \end{align*}
   Where $\tilde{\M{U}} = \phi (\M{X})(\otimes_{m=1}^M \M{U}_m)$.
\subsection{Proof for Proposition \ref{thm:proposition}}
\label{sec:derive-prop}
Consider a 3-mode  $T_1 \times T_2 \times T_3$ tensor $\W $  of functions $\W_{(1)} = [\V{w}_1(\M{X}), \cdots, \V{w}_T(\M{X})]$ 
\[ \W =  \T{S} \times_1 \M{U}_1(\T{X}) \times_2 \M{U}_2 \times_3 \M{U}_3 \]
where $\M{U}_m $ is an orthogonal $T_m \times R_m$ matrix. Assuming $\M{U}_1(\T{X})$ satisfies $\E [\M{U}_1^\top \M{U}_1] = \M{I}$ (orthogonal design after rotation).

With Tucker property
\[ \T{W}_{(1)} =  \M{U}_1(\T{X}) \T{S}_{(1)} (\M{U}_2 \M{U}_3)^\top \]

The population risk can be written as 
{\small
\[\loss(\T{W})  = \text{tr}\Big\{ ( \T{Y} - \langle \T{X} , \T{W}\rangle )( \T{Y} - \langle \T{X} , \T{W}\rangle )^\top \Big\}= \text{tr} \Big\{ \begin{pmatrix} 2\M{I} \\
- \T{S}_{(1)}(\M{U}_2 \M{U}_3)^\top \end{pmatrix}^\top \E[\text{cov}(\T{Y},   \M{U}_1(\T{X})]  \begin{pmatrix} \M{0} \\  -\T{S}_{(1)}(\M{U}_2 \M{U}_3)^\top  \end{pmatrix}+ \E(\T{Y}\T{Y}^\top) \Big\}  \]
}

Denote $\E[\text{cov}(\T{Y},   \M{U}_1(\T{X})] =  \M{\Sigma}(\M{U}_1)$, bound the difference
\begin{align*}
\loss(\W) - \hat{\loss}(\W) &= \text{tr}\Big\{   \begin{pmatrix} -2\M{I}\\ \T{S}_{(1)}(\M{U}_2 \M{U}_3)^\top\end{pmatrix}  (\M{\Sigma}(\M{U}_1) -   \hat{\M{\Sigma}}(\M{U}_1) )      \begin{pmatrix} \M{0} \\  \T{S}_{(1)}(\M{U}_2 \M{U}_3)^\top  \end{pmatrix}\Big\}\\
&\leq  \Big\|   \begin{pmatrix} -2\M{I}\\ \T{S}_{(1)}(\M{U}_2 \M{U}_3)^\top\end{pmatrix}  (\M{\Sigma}(\M{U}_1) -   \hat{\M{\Sigma}}(\M{U}_1) )    \Big\|_2   \Big\|   \begin{pmatrix} \M{0} \\  \T{S}_{(1)}(\M{U}_2 \M{U}_3)^\top  \end{pmatrix}  \Big \|_\star \\
&\leq C \max \{ 2,  \|  \T{S}_{(1)}  \|_\star^2\}  \|\M{\Sigma}(\M{U}_1) -   \hat{\M{\Sigma}}(\M{U}_1) \|_2 
\end{align*}
With $C$ as a universal constant. The inequality holds with   Schatten norm H\"{o}lder's inequality
\[ \|AB\|\\_{S_1} \leq \| A\|_{S_p}  \|B\|_{S_q} \quad 1/p + 1/q = 1\]
 
Given that $\text{sup}_{\M{U}_1} \|\M{\Sigma}(\M{U}_1) -   \hat{\M{\Sigma}}(\M{U}_1) \|_2  = \mathcal{O}_P \Big( \sqrt{\frac{T_2 T_3 + \log (T_1T_2 T_3)}{N}}\Big)$

Denote  empirical risk $\hat{\loss}  = \sum_{t =1}^T \sum_{i=1}^{n_t} \loss (\langle \V{w}_t, \V{x}_{t,i}\rangle $. Let $\W^\star = \text{inf}_{\W \in \T{C}} \loss(\W)$.  The excess risk
\begin{align*}
 \loss (\hat{\W})  - \loss (\W^\star)  &= \loss(\hat{\W}) - \hat{\loss}(\hat{\W})   +  ( \hat{\loss}(\hat{\W}) - \hat{\loss}(\W^\star  ) +  ( \hat{\loss}(\W^\star  - \loss (\W^\star)  )\\
 &\leq [ \loss(\hat{\W}) - \hat{\loss}(\hat{\W})]   -  [ \loss (\W^\star) - \hat{\loss}(\W^\star)  ]\\
 &\leq 2 \text{sup}_{\W \in \T{C}_N} \{ \loss(\W)   - \hat{\loss} (\W)\}\\
 &\leq  \mathcal{O}\Big(    \|\T{S}_{(1)} \| _\star^2 \|\M{\Sigma}(\M{U}_1) -   \hat{\M{\Sigma}}(\M{U}_1) \|_2     \Big)
\end{align*}
if we assume $\|\T{S}_{(1)}\|_\star^2 = \smallO( \Big(\frac{N}{T_2 T_3 + \log (T_1T_2 T_3)} \Big)^{1/4})$, then $ \loss (\hat{\W})  - \loss (\W^\star) \leq \smallO(1)$, thus we obtain the oracle inequality as stated.

\subsection{Proof of Theorem \ref{thm:theorem}}
\label{sec:derive-thm}
We can extend the approach of single task Gaussian process \citep{sollich2002learning} to our setting.  We provide the derivation for the full-rank case, but similar results apply to low-rank case as well. The Bayes error for the full-rank covariance model is:
\begin{equation*}
\hat{\epsilon} = \text{tr}(\M{\Lambda'}^{-1} + \M{\Psi}^\top \M{D}^{-1}\M{\Psi})^{-1}  
\end{equation*}
To obtain learning curve $\epsilon = \E_{\T{D}} [\hat{\epsilon}]$, it is useful to see how the matrix $\T{G}=(\M{\Lambda}^{-1} + \M{\Psi}^\top \M{D}^{-1}\M{\Psi})^{-1}$ changes with sample size.   $\M{\Psi}^\top\M{\Psi}$ can be interpreted as the input correlation matrix. 

To account for the fluctuations of the element in $\M{\Psi}^\top\M{\Psi}$, we introduce  auxiliary offset parameters $\{v_t \}$ into the definition of $\T{G}$.  Define resolvent matrix 
\[\T{G}^{-1}= \M{\Lambda}^{-1} + \M{\Psi}^\top \M{D}^{-1}\M{\Psi} +  \sum_t v_t \M{P}_t\]
where $\M{P}_t$ is short for $\M{P}_{t_1,\cdots,t_M}$, which defines the projection of $t$th task to its multi-directional indexes.

Evaluating the change
\[\T{G}(n+1)- \T{G}(n) = [\T{G}^{-1}(n) + \sigma_t^{-2}\psi_t \psi_t^\top]^{-1}- \T{G}(n) =\frac{\T{G}(n)\psi_t \psi_t^\top \T{G}(n)}{\sigma^2_t + \psi_t^\top \T{G}(n) \psi_t}\]
where element $(\psi_{t})_i = \delta_{\tau_{n+1}, t}\phi_{it}(x_{n+1})$ and $\tau$ maps the global sample index to task-specific sample index. Introducing $\M{G} = \E_{\T{D}}[ \T{G} ] $ and take expectation over numerator and denominator separately, we have
\[  \frac{\partial \M{G} }{\partial n_t}  = - \frac{\E_{\T{D}}[     \T{G}\M{P}_t\T{G}]}{\sigma^2_t   +   \text{tr} \M{P}_t \M{G}} \]

Since generalization error $\epsilon_t = \text{tr} \M{P}_t\M{G}$,  we have that $-\E_{\T{D}}[ \T{G}\M{P}_t \T{G} ] =  \frac{\partial}{\partial v_t} \E_{\T{D}}[ \T{G} ] = \frac{\partial \M{G}}{v_t}$. Multiplying $\M{P}_s$ on both sides yields the approximation for the expected change:
\[   \frac{\partial \M{P}_s\M{G}}{\partial n_t}   = \frac{\partial \epsilon_s}{\partial n_t}  = \frac{1}{\sigma_t^2 + \epsilon_t}  \frac{\partial \epsilon_s}{\partial v_t} \]

Solving $\epsilon_t(N,v)$ using the methods of characteristic curves and resetting $v$ to zero, gives the self-consistency equations:
\[    \epsilon_t(N) = \text{tr} \M{P}_t   \Big( \M{\Lambda'}^{-1} +\sum_s \frac{n_s}{\sigma^2_s + \epsilon_s  } \Big)^{-1}\]

}

\end{document}